\documentclass{article}

\usepackage{amsmath}
\usepackage{graphicx}
\PassOptionsToPackage{numbers,sort&compress}{natbib}
\usepackage[preprint]{neurips_2026}
\usepackage{multirow}
\usepackage{hyperref}
\usepackage{cleveref}
\usepackage{caption}
\usepackage{wrapfig}
\usepackage{array}
\usepackage{graphicx}
\usepackage{float}
\newcommand{\ACI}{A_{C\to I}}           
\newcommand{\AII}{A_{I\to I}}           

\newcommand{\Nimg}{N_\text{img}}                   

\usepackage[utf8]{inputenc} 
\usepackage[T1]{fontenc}    
\usepackage{hyperref}       
\usepackage{url}            
\usepackage{booktabs}       
\usepackage{amsfonts}       
\usepackage{nicefrac}       
\usepackage{microtype}      
\usepackage{xcolor}         

\title{AnchorDiff: Training-Free Concept Grounding for MM-DiTs via Anchor-Based Graph Propagation}

\author{%
  Jian Zhang\\
  School of Automation Science and Engineering \\
  South China University of Technology\\
  Guang Zhou, China \\
  \texttt{202411084134@mail.scut.edu.cn} \\
  \And
  Zhijun Zhang\\
  School of Automation Science and Engineering \\
  South China University of Technology\\
  Guang Zhou, China \\
  \texttt{auzjzhang@scut.edu.cn} \\
}

\begin{document}

\maketitle

\begin{abstract}
Multi-Modal Diffusion Transformers (MM-DiTs) encode rich representations for training-free concept grounding, but existing attention-based methods often produce overlapping activations on visually confusable concepts—a failure mode we call concept leakage—where target responses spill over to non-target objects. To address this issue, we propose \textbf{AnchorDiff}, a training-free grounding method that decouples semantic localization from structural refinement. AnchorDiff selects a high-confidence anchor from concept-to-image attention map and propagates it as a one-hot seed over a hybrid graph derived from image-to-image self-attention. The graph uses output-space similarity for dense within-object propagation and a row-wise attention gate to suppress cross-object connections. Additionally, we introduce the \textbf{Multi-Concept Confusion Dataset}, which contains images with multiple visually similar concepts and separate masks, enabling explicit evaluation of concept leakage. Experiments show that AnchorDiff achieves strong grounding performance on ImageNet-Segmentation and PascalVOC, while substantially reducing concept leakage on our Multi-Concept Confusion Dataset.
\end{abstract}

\vspace{-1em}
\section{Introduction}
\vspace{-0.5em}
\label{sec:intro}

Diffusion models have become powerful text-to-image generators~\citep{ho2020denoising,rombach2022high}, yet extracting spatially grounded semantic information from their internal representations remains challenging. Prior interpretability work has mainly studied UNet-based diffusion models, where cross-attention maps provide useful localization cues~\citep{tang2022daam,hertz2022prompt}. However, the interpretability of more recent diffusion transformers remains less explored, despite their growing adoption in state-of-the-art generative models \citep{peebles2023scalable}.

\begin{figure*}[t]
\centering
\includegraphics[width=0.95\textwidth]{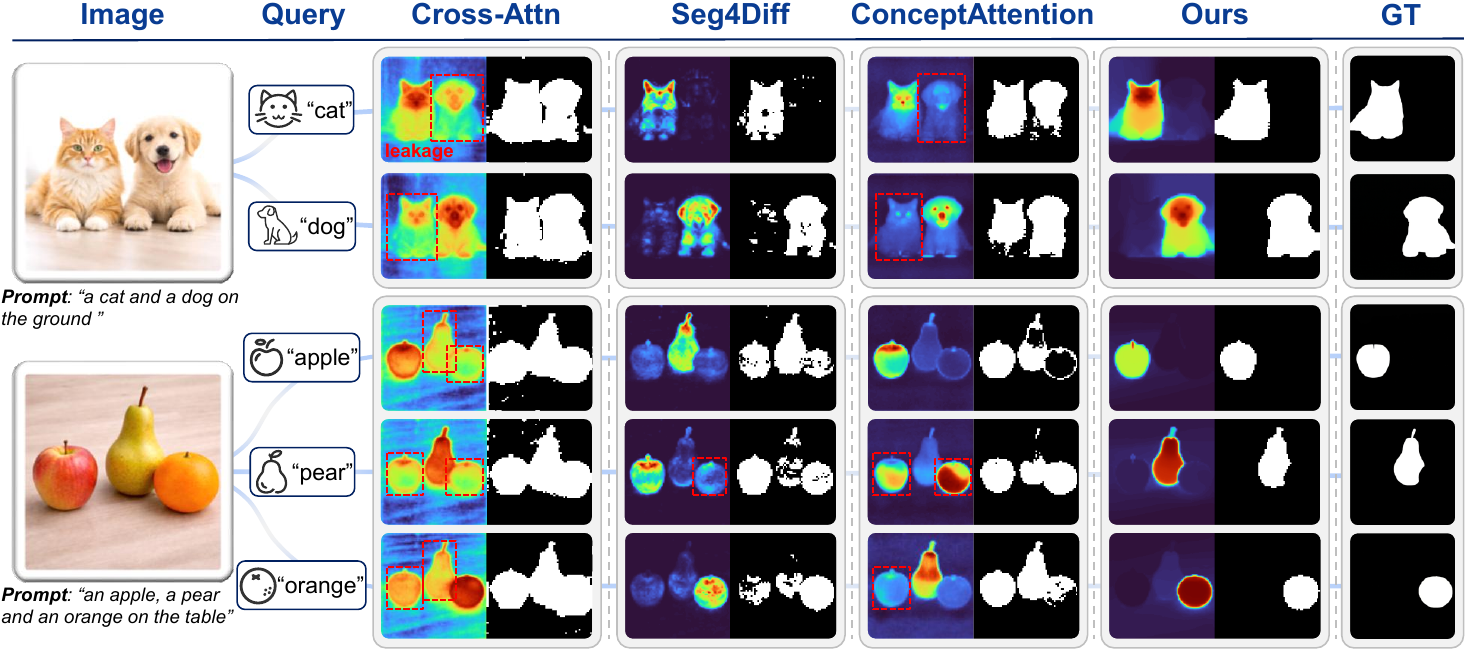}
\caption{\textbf{Concept leakage in semantic grounding.}
On visually similar concepts, existing methods suffer from concept leakage, where target responses spill over to non-target objects. AnchorDiff mitigates this by anchoring the target response and propagating it over an object-aware affinity graph.}
\label{fig:motivation}
\vspace{-1em}
\end{figure*}

Recent work has shown that DiT representations can support dense prediction and zero-shot segmentation~\citep{helbling2025conceptattention,kim2025seg4diff}, suggesting that their internal attention structures encode rich semantic and spatial information~\citep{esser2024scaling}. In particular, ConceptAttention~\citep{helbling2025conceptattention} computes output-space similarities between concept and image features and achieves strong zero-shot segmentation performance, in some cases surpassing CLIP-based methods~\citep{radford2021learning,chefer2021transformer,selvaraju2017grad,gandelsman2023interpreting,sun2024clip}.

Dense concept responses, obtained from concept-image similarity or concept-to-image attention, are often not sufficiently object-specific in scenes with visually similar concepts. In visually confusable scenes, dense responses overlap across multiple similar objects, causing target activation to spill over to non-target objects. We call this failure mode \emph{``concept leakage''}. As shown in \Cref{fig:motivation}, existing training-free MM-DiT methods such as raw cross-attention maps, Seg4Diff~\citep{kim2025seg4diff}, and ConceptAttention~\citep{helbling2025conceptattention} all exhibit varying degrees of concept leakage. ConceptAttention partially mitigates this by applying a pixelwise softmax over queried concepts, which improves contrast when distinct concepts are queried jointly, but the leakage among visually similar objects persists even under such competition. Moreover, with a single concept query, ConceptAttention relies on scene-dependent auxiliary background concepts such as ``grass'' or ``sky'' to produce a clear saliency map.

Our key observation is that image self-attention \citep{2017Attention} in Multi-modal Diffusion Transformers (MM-DiTs) provides structural cues that can separate visually confusable objects. Row-wise patterns of image-to-image attention $\AII$ reflect object boundaries: patches from different regions tend to attend differently. In contrast, output features $O_{I\to I}=A_{I\to I}V$ provide dense within-object connectivity, but can also connect visually similar objects. We therefore combine these two signals by using row-wise attention similarity as a structural gate on output-space affinity, yielding a graph that preserves within-object connectivity while suppressing cross-object propagation.

Based on this observation, we propose \textbf{AnchorDiff}, a training-free semantic grounding method for MM-DiT models. Instead of using dense concept responses directly, AnchorDiff first selects the maximum-response location of concept-to-image attention $\ACI$ as a high-confidence semantic anchor. It then initializes a one-hot seed at this anchor and propagates it over the hybrid self-attention graph. This decouples semantic localization from structural refinement: $\ACI$ identifies where the target concept is likely to appear, while the $\AII$-derived graph determines how activation should propagate within object boundaries. To explicitly evaluate concept leakage, we introduce the \textbf{Multi-Concept Confusion Dataset}, a benchmark of images with two or three visually similar foreground concepts and separate per-concept masks. Unlike standard foreground segmentation datasets, it evaluates whether a method can localize the queried concept while suppressing activation on similar non-target objects.

In summary, our contributions are:
\vspace{-0.5em}
\begin{itemize}
\setlength{\itemsep}{0.2em}
\setlength{\topsep}{0.2em}
\setlength{\parsep}{0pt}
\setlength{\parskip}{0pt}
\item We identify {concept leakage} as a key limitation of attention-based semantic grounding in visually confusable scenes. To address it, we propose {AnchorDiff}, a training-free framework that combines $\ACI$-based sparse semantic anchoring with $\AII$-based graph propagation for object-aware refinement.

\item We show that MM-DiT image self-attention provides complementary cues for graph construction: row-wise attention similarity suppresses cross-object connections, while output-space affinity derived from self-attention outputs provides dense within-object connectivity.

\item We introduce the {Multi-Concept Confusion Dataset}, a dedicated benchmark for evaluating concept leakage, containing images with multiple visually similar concepts and separate per-concept masks.

\item We demonstrate that {AnchorDiff} achieves state-of-the-art training-free grounding performance on ImageNet-Segmentation, PascalVOC, and our Multi-Concept Confusion Dataset, improving localization quality while substantially reducing non-target activation.
\end{itemize}

\vspace{-0.5em}
\section{Related work}
\vspace{-0.5em}
\label{sec:related}

\paragraph{UNet-based diffusion grounding.}
Cross-attention maps in UNet-based diffusion models provide spatial localization of text concepts. DAAM~\citep{tang2022daam} aggregates cross-attention across timesteps to produce word-level heatmaps. DiffSegmenter~\citep{wang2025diffusion} segments Stable Diffusion outputs by multiplying a dense cross-attention score map with a raw self-attention map in a single step. OVAM~\citep{marcos2024open} introduces token-optimized attention maps for open-vocabulary segmentation in SDXL~\citep{podellsdxl}. These methods operate on UNet-based diffusion models with separate image and text processing streams, whereas AnchorDiff targets MM-DiT models with joint image-text attention, enabling the extraction of cross-modal and intra-modal attention signals.

\vspace{-0.5em}
\paragraph{DiT-based semantic grounding.}
Recent work has begun to repurpose DiT representations for segmentation and semantic grounding. ConceptAttention~\citep{helbling2025conceptattention} injects concept tokens as attention probes and computes output-space concept-image similarities with a pixelwise softmax over concepts. Seg4Diff~\citep{kim2025seg4diff} analyzes MM-DiT attention and identifies expert layers whose cross-modal attention aligns text tokens with spatially coherent image regions. While these methods reveal strong semantic localization signals in DiT representations, they obtain masks from dense output-space concept responses or dense cross-modal attention maps. Such dense responses can overlap in visually confusable scenes, leading to concept leakage. AnchorDiff instead uses concept-to-image attention only to select a sparse semantic anchor, and then propagates it through an object-aware graph derived from image self-attention.

\begin{figure*}[t]
\centering
\includegraphics[width=.95\textwidth]{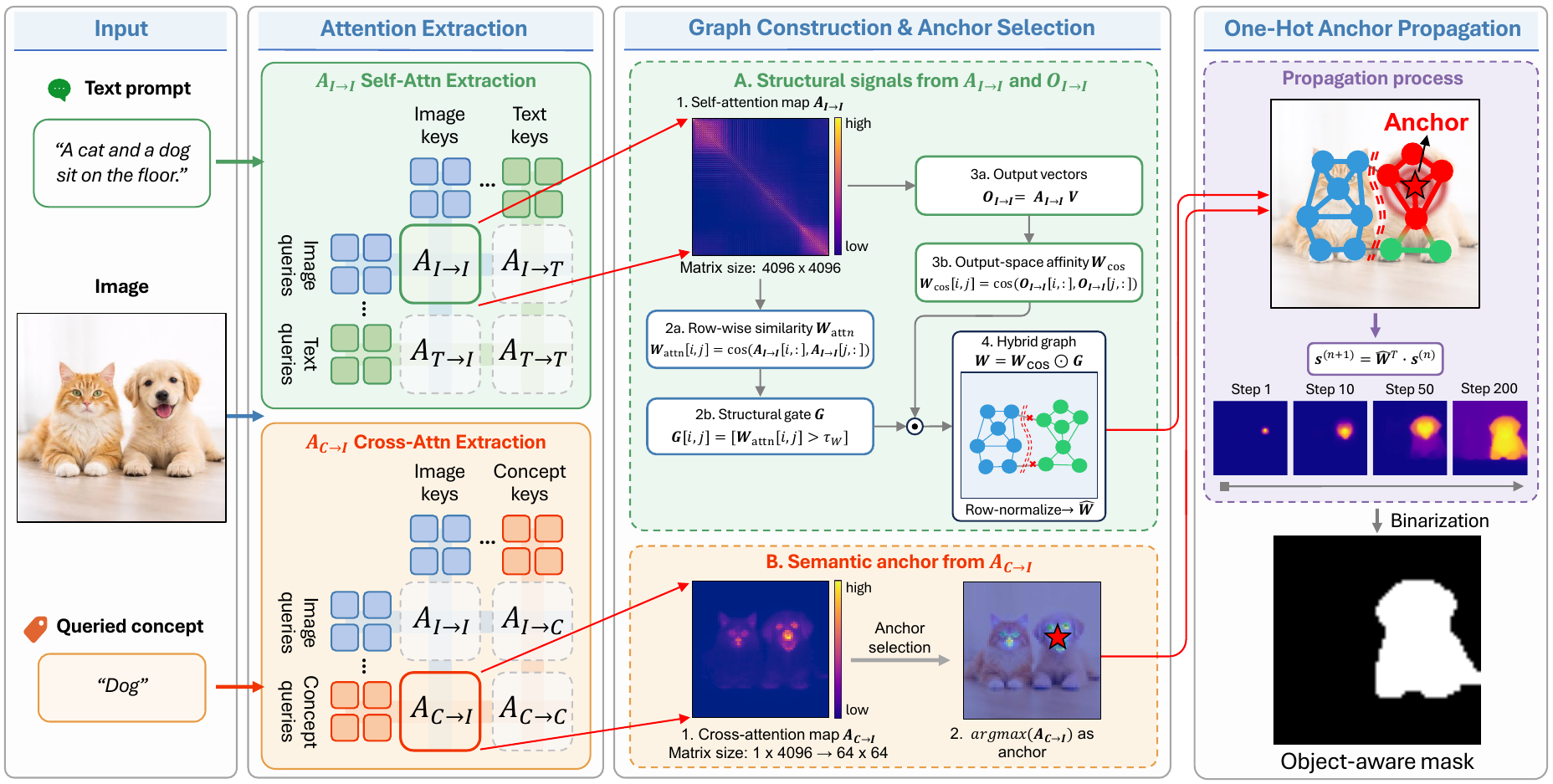}
\caption{\textbf{Overview of AnchorDiff.}
AnchorDiff extracts concept-to-image attention for anchor selection and image-to-image self-attention for graph construction. A one-hot seed at the semantic anchor is propagated over a hybrid graph combining output-space affinity with a row-wise attention gate, producing object-confined masks with reduced concept leakage.}
\label{fig:method_overview}
\vspace{-1em}
\end{figure*}

\vspace{-0.5em}
\section{Preliminaries}
\vspace{-0.5em}

Before introducing AnchorDiff, we analyze the image-to-image self-attention of MM-DiT. Raw attention weights are too local for direct propagation, but row-wise attention patterns capture object boundaries, while self-attention output-space features provide within-object connectivity.

\vspace{-0.5em}
\subsection{Locality of raw image-to-image attention}
\vspace{-0.5em}

\begin{figure*}[t]
\centering
\begin{minipage}[t]{0.48\textwidth}
  \centering
  \includegraphics[width=\linewidth]{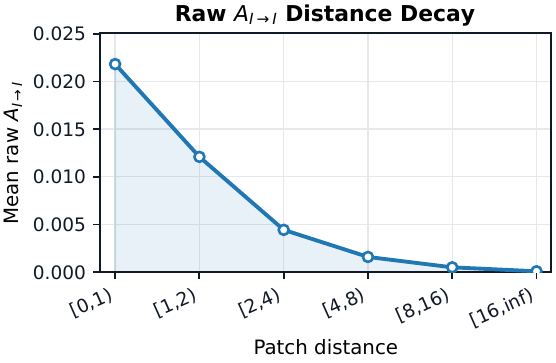}
  \captionof{figure}{
  \textbf{Raw image-to-image attention is spatially local.}
  Mean raw attention weights are measured over patch-pair distance bins and decay rapidly on the $64\times64$ grid.
  }
  \label{fig:property1_locality}
\end{minipage}
\hfill
\begin{minipage}[t]{0.48\textwidth}
  \centering
  \includegraphics[width=\linewidth]{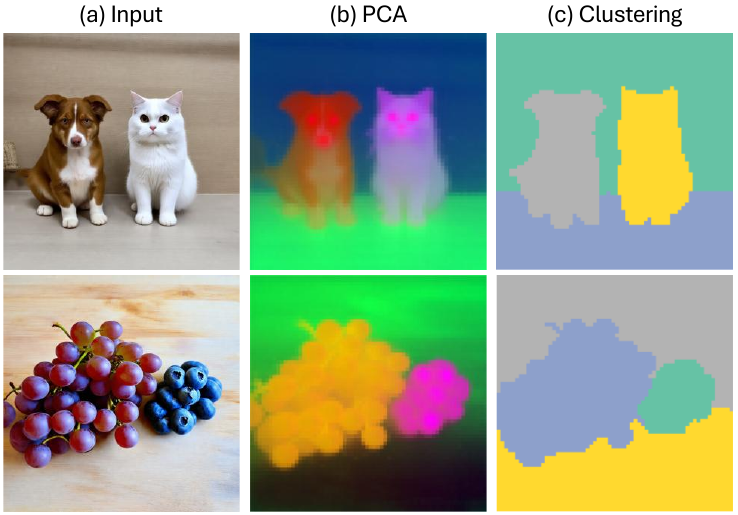}
  \captionof{figure}{
  \textbf{Row-wise consistency in image self-attention.}
  PCA projections and clustering of $A_{I\to I}$ rows form spatially coherent object regions, without using concept labels.
  }
  \label{fig:property2}
\end{minipage}
\vspace{-1em}
\end{figure*}

Let $A_{I\to I}\in\mathbb{R}^{N_{\mathrm{img}}\times N_{\mathrm{img}}}$ denote the image-to-image self-attention matrix extracted from MM-DiT. Although $A_{I\to I}$ captures interactions among image tokens, its weights are concentrated around local neighborhoods. On the Multi-Concept Confusion Dataset, the mean attention weight drops from $2.18\times10^{-2}$ for neighboring patches to $8.6\times10^{-5}$ for patches more than 16 tokens apart, as shown in \Cref{fig:property1_locality}. Thus, raw $A_{I\to I}$ alone provides insufficient long-range connectivity for object propagation.

\vspace{-0.5em}
\subsection{Object structure from row-wise attention similarity}
\vspace{-0.5em}

Although individual entries of $A_{I\to I}$ are dominated by local neighbors, each row $A_{I\to I}[i,:]$ summarizes how patch $i$ relates to all image tokens. Patches from the same object tend to have similar row patterns, while patches across object boundaries differ. As shown in \Cref{fig:property2}, PCA projections and clustering of attention rows form spatially coherent object regions without using concept labels. We therefore define row-wise structural similarity as
\begin{equation}
W_{\mathrm{attn}}[i,j]
=
\cos\!\left(A_{I\to I}[i,:], A_{I\to I}[j,:]\right).
\end{equation}
On the Multi-Concept Confusion Dataset, $W_{\mathrm{attn}}$ is higher for same-object pairs ($21.83\%$) than for confusable different-object pairs ($4.65\%$) or foreground-background pairs ($1.13\%$) (\Cref{tab:property3_affinity_gating}), indicating that it provides a concept-agnostic cue for suppressing cross-object connections.

\begin{wraptable}{r}{0.47\textwidth}
\vspace{-1.2em}
\centering
\scriptsize
\renewcommand{\arraystretch}{0.9}
\caption{
\textbf{Affinity statistics.}
Mean affinities (\%) for same-object, confusable different-object, and foreground-background patch pairs.
}
\label{tab:property3_affinity_gating}
\begin{tabular}{>{\arraybackslash}p{1.8cm}|%
>{\centering\arraybackslash}p{1cm}%
>{\centering\arraybackslash}p{1.1cm}%
>{\centering\arraybackslash}p{1cm}}
\toprule
Affinity & Same obj. & Conf. diff. & FG-BG \\
\midrule
$W_{\mathrm{attn}}$ & 21.83 & 4.65 & 1.13 \\
$W_{\mathrm{cos}}$ & 79.40 & 67.27 & 41.80 \\
$W_{\mathrm{cos}}\odot G_{\mathrm{attn}}$ & 10.14 & 0.04 & 0.03 \\
\bottomrule
\end{tabular}

\vspace{0.7em}
\renewcommand{\arraystretch}{0.9}
\caption{
\textbf{Anchor localization accuracy.}
The selected anchor is correct if it lies inside the ground-truth mask of the queried concept.
}
\label{tab:anchor_hit_rate}
\vspace{-0.4em}
\begin{tabular}{>{\arraybackslash}p{1.8cm}|%
>{\centering\arraybackslash}p{1cm}%
>{\centering\arraybackslash}p{1.1cm}%
>{\centering\arraybackslash}p{1cm}}
\toprule
Model & Correct & Total & Rate (\%) \\
\midrule
SD3 & 2532 & 2842 & 89.09 \\
SD3.5 & 2565 & 2842 & 90.25 \\
\bottomrule
\end{tabular}

\vspace{-0.5em}
\end{wraptable}

\vspace{-0.5em}
\subsection{Output-space affinity for dense propagation}
\vspace{-0.5em}

\begin{figure*}[t]
\centering
\includegraphics[width=0.95\textwidth,height=0.19\textheight]{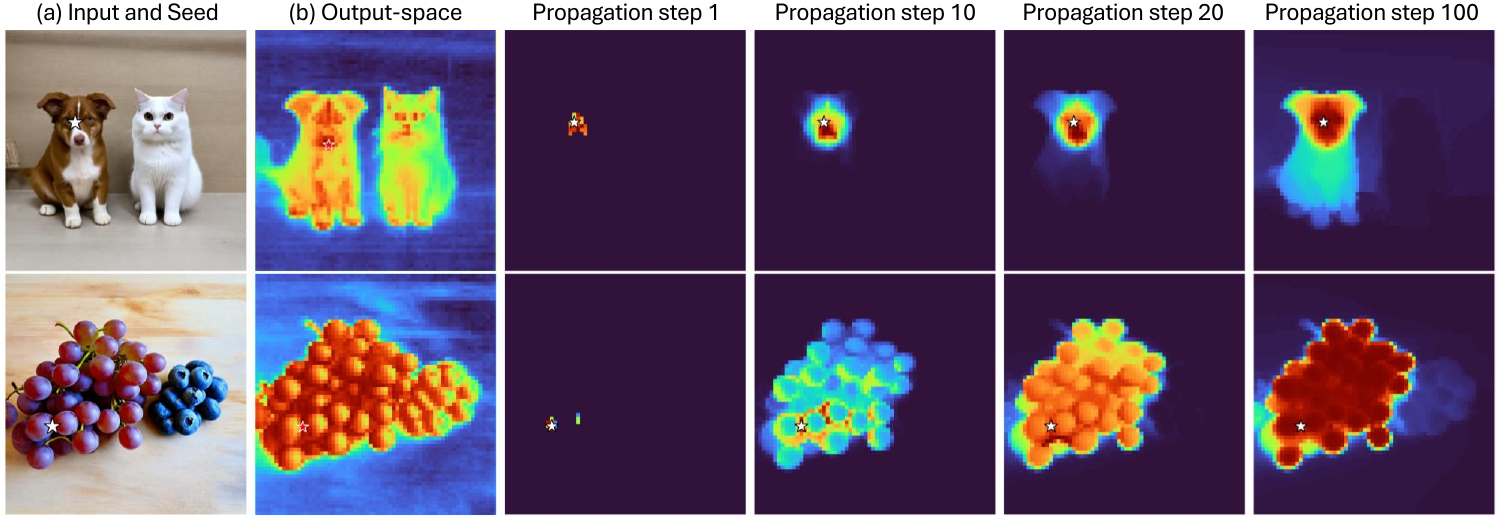}
\caption{
\textbf{Output-space propagation with structural gating.}
Output-space affinity yields dense same-object responses but also activates
similar non-target objects. After gating with $W_{\mathrm{attn}}$, propagation from a one-hot anchor recovers the target object while reducing cross-object leakage.
}
\label{fig:property3}
\vspace{-1em}
\end{figure*}

MM-DiT self-attention also produces output-space features
$O_{I\to I} = A_{I\to I} V$, where $V$ denotes the Value projection of image tokens. We define the output-space affinity as:
\begin{equation}
W_{\mathrm{cos}}[i,j]=\cos\!\left(O_{I\to I}[i,:],\, O_{I\to I}[j,:]\right).
\end{equation}
Compared with raw attention weights or row-wise attention similarity alone, $W_{\mathrm{cos}}$ provides denser long-range connectivity among the same-object patches, making it suitable for propagating activation across the entire object. 
As shown in \Cref{fig:property3}, output-space similarity produces dense responses over same-object regions but also activates visually similar non-target objects, causing cross-object leakage. The statistics in \Cref{tab:property3_affinity_gating} further quantify this trade-off. $W_{\mathrm{cos}}$ assigns high affinity to same-object pairs ($79.40\%$), but also to confusable different-object pairs ($67.27\%$) and foreground-background pairs ($41.80\%$). In contrast, gating with $W_{\mathrm{attn}}$ reduces cross-object and foreground-background affinities to $0.04\%$ and $0.03\%$. We therefore use $W_{\mathrm{cos}}$ as the propagation weight and thresholded $W_{\mathrm{attn}}$ as the structural gate.


\vspace{-0.5em}
\section{Method}
\vspace{-0.5em}

We propose {AnchorDiff}, a training-free grounding method for MM-DiT models. As shown in \Cref{fig:method_overview}, AnchorDiff decouples semantic localization from structural refinement: it selects a high-confidence anchor from concept-to-image attention map $A_{C\to I}$ and propagates a one-hot seed over a hybrid affinity graph constructed from image self-attention signals $A_{I\to I}$ and $O_{I\to I}$.

\vspace{-0.5em}
\subsection{Attention extraction}
\vspace{-0.5em}

Given a text prompt and target concepts $C=\{c_1,\dots,c_K\}$, we extract image-to-image self-attention for structural propagation and concept-to-image attention for semantic anchor selection. From the standard image-text denoising process, we record
\begin{equation}
A_{I\to I}^{(\ell)}
=
\mathrm{softmax}\!\left(
\frac{
Q_{\mathrm{img}}^{(\ell)}
\left(K_{\mathrm{img}}^{(\ell)}\right)^\top
}{\sqrt{d_h}}
\right)
\in \mathbb{R}^{N_{\mathrm{img}}\times N_{\mathrm{img}}},
\end{equation}
where $\ell$ indexes the transformer layer, $N_{\mathrm{img}}$ is the number of image tokens, and $d_h$ is the attention head dimension. We also record the corresponding self-attention output features
\begin{equation}
O_{I\to I}^{(\ell)}
=
A_{I\to I}^{(\ell)}V_{\mathrm{img}}^{(\ell)}
\in \mathbb{R}^{N_{\mathrm{img}}\times d_h}.
\end{equation}

We then perform a second forward pass with the target concept tokens injected following~\citep{helbling2025conceptattention} and extract concept-to-image attention
\begin{equation}
A_{C\to I}^{(\ell)}
=
\mathrm{softmax}\!\left(
\frac{
Q_{\mathrm{concept}}^{(\ell)}
\left(K_{\mathrm{img}}^{(\ell)}\right)^\top
}{\sqrt{d_h}}
\right)
\in \mathbb{R}^{K\times N_{\mathrm{img}}},
\end{equation}
where each row gives a coarse response for one queried concept. Finally, we aggregate the extracted attention maps and output features over selected layers for anchor selection and graph construction:
\begin{equation}
 A_{C\to I}
=
\frac{1}{|L|}\sum_{\ell\in L} A_{C\to I}^{(\ell)},
\quad
 A_{I\to I}
=
\frac{1}{|L|}\sum_{\ell\in L} A_{I\to I}^{(\ell)},
\quad
 O_{I\to I}
=
\frac{1}{|L|}\sum_{\ell\in L} O_{I\to I}^{(\ell)}.
\end{equation}

\vspace{-0.5em}
\subsection{Semantic anchor selection}
\vspace{-0.5em}

Dense concept responses usually overlap in visually confusable scenes. Instead of using the full response map from $A_{C\to I}$, AnchorDiff retains only the maximum-response location for each concept. For concept $c_k$, we define
\begin{equation}
h_{c_k} =  A_{C\to I}[k,:],\quad
p^\ast_k = \arg\max_i h_{c_k}[i].
\end{equation}
We use $p^\ast_k$ as a semantic anchor and initialize a one-hot seed at this location for graph propagation. To verify the reliability of this anchor, we measure whether $p^\ast_k$ falls inside the ground-truth mask of the queried concept $c_k$. On the Multi-Concept Confusion Dataset, the anchor localization accuracy reaches $89.09\%$ for SD3 and
$90.25\%$ for SD3.5-Large over 2842 image-concept pairs, as shown in
\Cref{tab:anchor_hit_rate}. This indicates that the peak of $A_{C\to I}$ usually provides a reliable seed for graph propagation.

\vspace{-0.5em}
\subsection{Hybrid affinity graph construction}
\vspace{-0.5em}

We construct a hybrid affinity graph from $ A_{I\to I}$ and $ O_{I\to I}$. We first compute row-wise attention similarity:
\begin{equation}
W_{\mathrm{attn}}[i,j]
=
\frac{
 A_{I \to I}[i,:] \cdot  A_{I \to I}[j,:]
}{
\left\lVert  A_{I \to I}[i,:] \right\rVert
\left\lVert  A_{I \to I}[j,:] \right\rVert
}.
\end{equation}

We binarize $W_{\mathrm{attn}}$ into a structural gate by retaining only the top $2\%$ entries:
\begin{equation}
G[i,j]
=
\mathbb{1}\left[
W_{\mathrm{attn}}[i,j] > \tau_W
\right],
\end{equation}
where $\tau_W$ is the $98$th percentile of $W_{\mathrm{attn}}$ entries. We then compute output-space affinity from $ O_{I\to I}$:
\begin{equation}
W_{\mathrm{cos}}[i,j]
=
\frac{
 O_{I \to I}[i,:] \cdot  O_{I \to I}[j,:]
}{
\left\lVert  O_{I \to I}[i,:] \right\rVert
\left\lVert  O_{I \to I}[j,:] \right\rVert
}.
\end{equation}
The final graph uses $W_{\mathrm{cos}}$ as propagation weights and $G$ as the structural gate:
\begin{equation}
W[i,j] = W_{\mathrm{cos}}[i,j]\cdot G[i,j],
\quad
\hat W[i,:] = \frac{W[i,:]}{\sum_j W[i,j]}.
\end{equation}

\vspace{-1em}
\subsection{One-hot anchor propagation}
\vspace{-0.5em}

We initialize a one-hot seed $s^{(0)}\in\mathbb{R}^{N_{\mathrm{img}}}$ at the anchor $p_k^\ast$, with $s^{(0)}_{p_k^\ast}=1$ and all other entries set to 0. We then propagate it over the normalized affinity graph:
\begin{equation}
s^{(n+1)}=\hat W^\top s^{(n)},
\quad n=0,\dots,N_{\mathrm{diff}}-1.
\end{equation}
The structural gate restricts propagation to structurally consistent high-affinity regions. After $N_{\mathrm{diff}}$ steps, we obtain
\begin{equation}
\hat h_{c_k} = \mathrm{minmax}\!\bigl(s^{(N_{\mathrm{diff}})}\bigr),
\end{equation}
where $\mathrm{minmax}(\cdot)$ linearly normalizes the response to $[0,1]$. We obtain the final binary mask by thresholding $\hat h_{c_k}$ at its mean value.

\vspace{-0.5em}
\section{Experiments}
\vspace{-0.5em}
\label{sec:experiments}

We evaluate AnchorDiff in two settings: single-concept grounding on standard benchmarks, and multi-concept grounding under visually similar distractors using our Multi-Concept Confusion Dataset.

\vspace{-0.5em}
\subsection{Experimental setup}
\vspace{-0.5em}
\label{sec:exp:setup}

\begin{figure*}[t]
\centering
\includegraphics[width=0.95\textwidth]{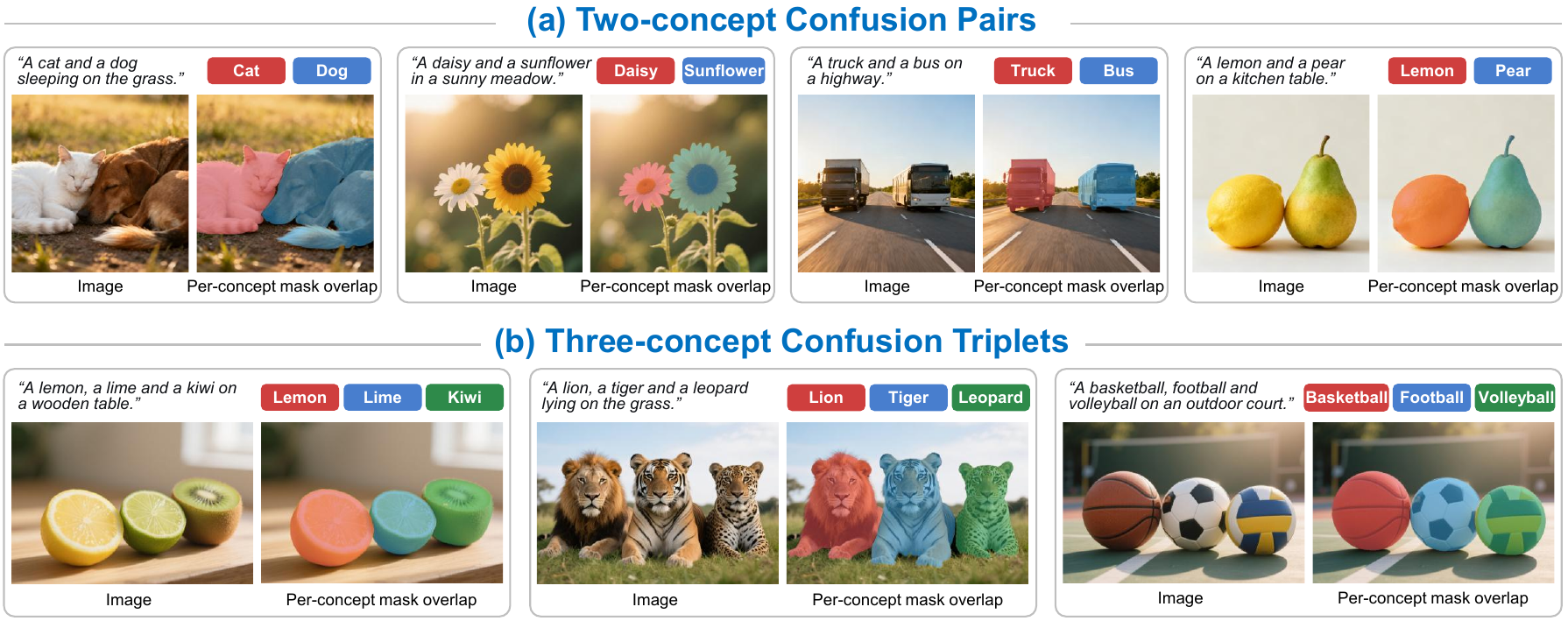}
\caption{
Representative samples from our Multi-Concept Confusion Dataset. Each image contains two or three visually similar concepts, with per-concept mask overlaps shown in different colors.
}
\label{fig:dataset}
\vspace{-1em}
\end{figure*}

\paragraph{Models.}
We apply AnchorDiff to two frozen MM-DiT backbones, {SD3 Medium} and {SD3.5 Large}~\citep{esser2024scaling}. All experiments are training-free.

\vspace{-0.5em}
\paragraph{Implementation details.}
All images are resized to $1024 \times 1024$, corresponding to a $64 \times 64$ latent grid with $\Nimg=4096$ image tokens. AnchorDiff extracts masks from MM-DiT attention at the midpoint of the generation trajectory. For existing image benchmarks, we apply the same midpoint extraction along the inversion trajectory. This single-timestep extraction avoids full-trajectory attention aggregation and keeps inference efficient. For semantic anchoring, we aggregate $A_{C\to I}$ over all transformer layers to choose the target-concept anchor, since anchor selection only requires the $\arg\max$ location and benefits from broad semantic averaging. For graph construction, we use image-to-image attention layers $\{9,18\}$ for SD3 and $\{10,23\}$ for SD3.5, as graph propagation is sensitive to layer-specific structural quality (\Cref{fig:layer_sensitivity}). In experiments, we set the structural gate threshold to $\tau_W=0.98$ and use $N_{\mathrm{diff}}=160$ propagation steps for both backbones. All experiments are run on NVIDIA GeForce RTX 4090 GPUs. AnchorDiff takes approximately 0.52 seconds per concept query on a single GPU, covering attention extraction and 160-step graph propagation.

\vspace{-0.5em}
\paragraph{Datasets.}
For single-concept grounding, we evaluate on {ImageNet-Segmentation}~\citep{guillaumin2014imagenet} and {PascalVOC 2012 Single-Class}~\citep{2015The}, where the latter is constructed by selecting PascalVOC 2012 validation images with exactly one annotated foreground class. To evaluate concept leakage, we construct a {Multi-Concept Confusion Dataset} with 1,272 images containing two or three visually similar foreground concepts, each with a separate mask. The dataset includes synthetic images from FLUX.1-dev and Seedream 3.0/4.0~\citep{gao2025seedream,seedream2025seedream}, as well as real web images. We use Qwen2.5-VL~\citep{bai2025qwen3} to localize each concept and SAM2~\citep{ravisam} to segment the corresponding masks, followed by manual inspection. Representative examples are shown in \Cref{fig:dataset}.

\begin{table*}[t]
\centering
\renewcommand{\arraystretch}{0.88}
\caption{Main results on standard zero-shot segmentation benchmarks. We report Acc, mIoU, and mAP on ImageNet-Segmentation and PascalVOC Single-Class.}
\scriptsize
\setlength{\tabcolsep}{4pt} 
\begin{tabular}{ p{3.6cm} | 
  >{\centering\arraybackslash}p{1.4cm} 
  >{\centering\arraybackslash}p{1.4cm} 
  >{\centering\arraybackslash}p{1.4cm} | 
  >{\centering\arraybackslash}p{1.4cm} 
  >{\centering\arraybackslash}p{1.4cm} 
  >{\centering\arraybackslash}p{1.4cm}}
\toprule
\multirow{2}{*}{\textbf{Method}} 
& \multicolumn{3}{c|}{\textbf{ImageNet-Segmentation}} 
& \multicolumn{3}{c}{\textbf{PascalVOC Single-Class}}\\
& Acc $\uparrow$ & mIoU $\uparrow$ & mAP $\uparrow$ 
& Acc $\uparrow$ & mIoU $\uparrow$ & mAP $\uparrow$\\

\midrule
LRP~\citep{binder2016layer}                & 50.55 & 33.82 & 55.92 & 48.02 & 30.71 & 53.47 \\
Partial-LRP~\citep{binder2016layer}        & 75.23 & 60.29 & 83.21 & 71.50 & 51.38 & 84.37 \\
Rollout~\citep{abnar2005quantifying}            & 70.51 & 54.46 & 81.73 & 69.81 & 51.25 & 85.33 \\
ViT Attention~\citep{2020An}      & 70.62 & 54.58 & 78.86 & 68.51 & 44.81 & 83.63 \\
GradCAM~\citep{selvaraju2017grad}            & 68.44 & 52.02 & 70.68 & 70.50 & 44.97 & 76.92 \\
TextSpan~\citep{gandelsman2023interpreting}           & 75.81 & 61.05 & 80.86 & 75.00 & 56.24 & 84.81 \\
TransInterp~\citep{chefer2021transformer}        & 78.43 & 64.51 & 84.36 & 76.98 & 57.24 & 86.77 \\
CLIPasRNN~\citep{sun2024clip}          & 83.05  & 71.01 & 86.59 & 83.44 & 70.79 &  90.98 \\
DINO SA~\citep{caron2021emerging}            & 81.76 & 69.14 & 86.05 & 79.51 & 62.62 & 88.53 \\
DINOv2 SA~\citep{oquab2024dinov2}          & 83.20 & 71.24 & 85.19 & 81.72 & 64.76 & 87.94 \\
DINOv2 Reg SA~\citep{darcet2024vision}      & {84.21} & 73.24 & 86.62 & 83.36 & 67.27 &  88.97 \\
iBOT SA~\citep{zhou2021ibot}            & 83.18 & 71.21 & 86.51 & 80.91  & 64.95  &  89.07 \\
DAAM SDXL~\citep{tang2022daam}          &82.33  &69.97  &{90.48}  &74.63  &58.18  & 88.79 \\
DAAM SD2~\citep{tang2022daam}           &83.78  &72.09  &88.95  &78.80  &63.29  & 89.55  \\
OVAM~\citep{marcos2024open}          & 83.88  & 72.24 & 89.65 & 80.40 & 65.71 &  91.92 \\
Cross Attention SD3   &80.90  &67.92  &86.79  &82.92  &68.77  & 92.16 \\
Cross Attention SD3.5   &76.52 &61.97 &89.37  &74.23  & 58.01  & 92.44  \\
Seg4Diff SD3~\citep{kim2025seg4diff}        & 82.13  & 69.68 & 83.84 & 83.84 & 69.96 & 87.28\\
Seg4Diff SD3.5~\citep{kim2025seg4diff}      &79.53 &66.01 & 83.29 & 82.97 & 67.88 & 87.36 \\
ConceptAttention SD3~\citep{helbling2025conceptattention}  &81.48  &68.75  &85.52  &85.77  & 72.04  &88.48   \\
ConceptAttention SD3.5~\citep{helbling2025conceptattention} &83.75  &72.04  &88.48  & 85.11 & 71.56  & 92.24 \\
\midrule
\textbf{Ours SD3}      &83.76  &{75.09}  &90.06  & {87.30}  & {79.74}  & {93.15} \\
\textbf{Ours SD3.5}      & \textbf{85.46}   & \textbf{77.29}  & \textbf{91.20} & \textbf{88.50} & \textbf{81.53} & \textbf{94.05} \\
\bottomrule
\end{tabular}
\label{tab:main_results_standard}
\vspace{-1em}
\end{table*}

\vspace{-0.5em}
\paragraph{Metrics.}
For ImageNet-Segmentation and PascalVOC Single-Class, we follow the standard binary foreground segmentation protocol and report Acc, mIoU, and mAP over the whole image, treating the queried class as foreground and all remaining pixels as background. For the Multi-Concept Confusion Dataset, we evaluate each image once per target concept. Given a target mask $M_c$, we define $M_{\mathrm{other}}$ as the union of all non-target concept masks, excluding pixels overlapping with $M_c$. Unlike the standard metrics, $\mathrm{mIoU}_{fg}$, $\mathrm{mAP}_{fg}$, and $\mathrm{Acc}_{fg}$ are computed only on the annotated foreground region $M_c \cup M_{\mathrm{other}}$, measuring whether the queried target concept is separated from visually similar non-target concepts without involving background pixels. We also report the Non-target Activation Ratio (NAR):
\begin{equation}
\mathrm{NAR}
=
\frac{
\sum_{p \in M_{\mathrm{other}}} H_c(p)
}{
\sum_{p \in M_c \cup M_{\mathrm{other}}} H_c(p)
},
\end{equation}
where $H_c$ is the response map for concept $c$. Lower NAR indicates less activation leakage to visually similar non-target objects. All multi-concept metrics are averaged over image-concept pairs.

\vspace{-0.5em}
\paragraph{Baselines.}
We compare with three groups of baselines: CLIP/ViT interpretability methods, including LRP~\citep{binder2016layer}, Partial-LRP~\citep{binder2016layer}, Rollout~\citep{abnar2005quantifying}, GradCAM~\citep{selvaraju2017grad}, ViT Attention~\citep{2020An}, TextSpan~\citep{gandelsman2023interpreting}, TransInterp~\citep{chefer2021transformer}, and CLIPasRNN~\citep{sun2024clip}; self-supervised ViT self-attention baselines, including DINO SA~\citep{caron2021emerging}, DINOv2 SA~\citep{oquab2024dinov2}, DINOv2 Reg SA~\citep{darcet2024vision}, and iBOT SA~\citep{zhou2021ibot}; and diffusion-based grounding methods, including DAAM~\citep{tang2022daam}, OVAM~\citep{marcos2024open}, Seg4Diff~\citep{kim2025seg4diff}, ConceptAttention~\citep{helbling2025conceptattention}, and raw SD3/SD3.5 cross-attention maps. For the Multi-Concept Confusion Dataset, we report only methods that can generate a separate response map for each queried concept.

\vspace{-0.5em}
\subsection{Main results}
\vspace{-0.5em}
\label{sec:exp:main}

\paragraph{Concept leakage evaluation.}
\Cref{tab:multi_concept_results} evaluates whether methods can localize a queried concept while suppressing visually similar non-target concepts. Ours SD3.5 achieves 68.85\% mIoU$_{fg}$ and 89.97\% Acc$_{fg}$, outperforming Seg4Diff SD3.5 by 22.28 mIoU points and 20.78 accuracy points. It also reduces NAR to 12.21, compared with 19.08 for Seg4Diff SD3.5 and 26.36 for ConceptAttention SD3.5, indicating substantially less activation leakage. These results show that dense semantic responses are prone to concept leakage, while AnchorDiff produces more object-confined masks by propagating a sparse semantic anchor over a structurally gated graph.

\begin{figure}[t]
\centering
\includegraphics[width=0.95\linewidth, height=0.263\textheight]{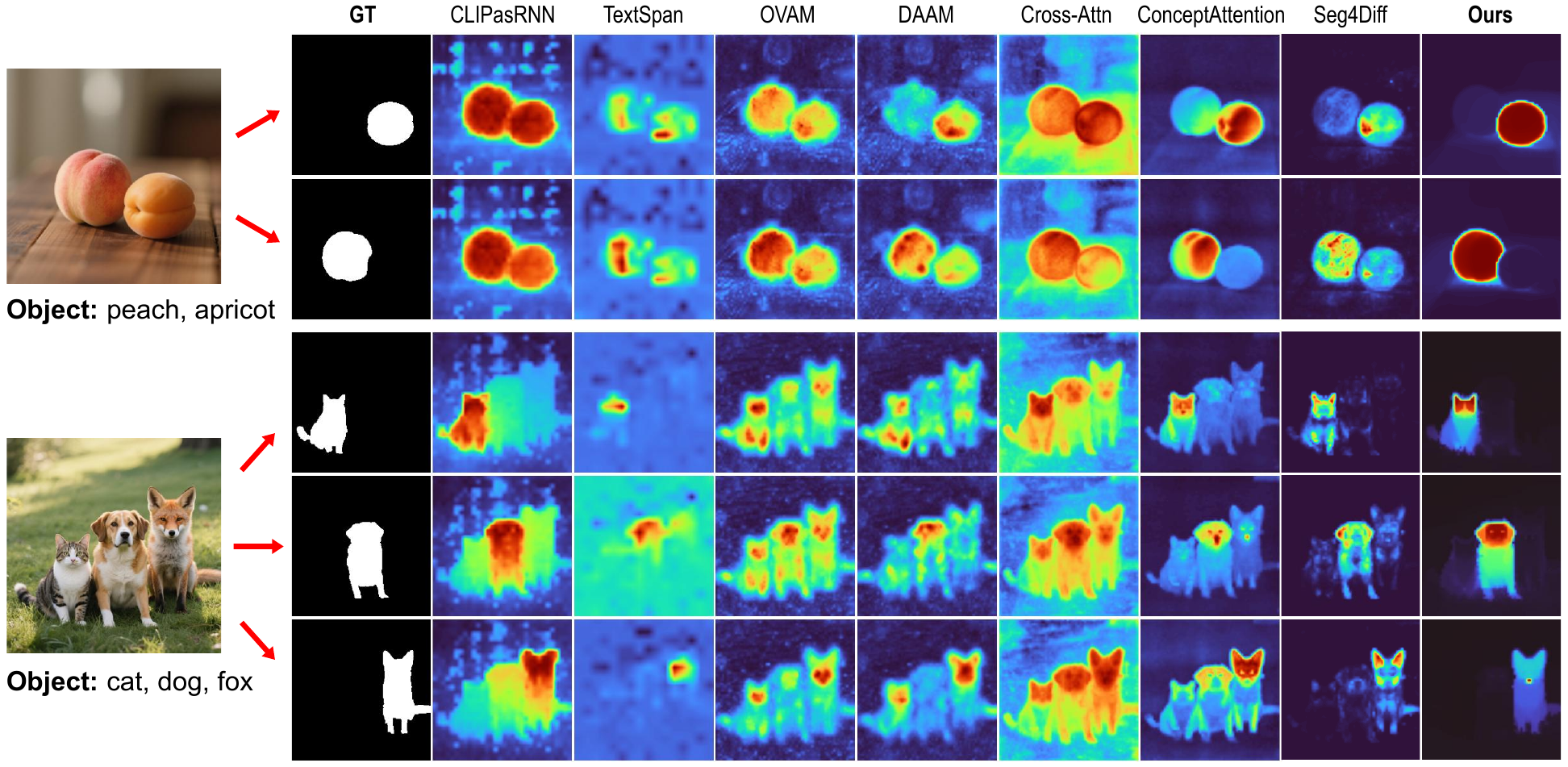}
\caption{
Qualitative results on representative samples from the Multi-Concept Confusion Dataset.
}
\label{fig:qualitative_multiconcept}
\vspace{-1em}
\end{figure}

\vspace{-0.5em}
\paragraph{Standard single-concept grounding.}
\Cref{tab:main_results_standard} reports results on ImageNet-Segmentation and PascalVOC Single-Class. AnchorDiff consistently outperforms all compared baselines on both benchmarks. Ours SD3.5 achieves 77.29\% mIoU on ImageNet-Segmentation and 81.53\% mIoU on PascalVOC Single-Class, surpassing prior diffusion-based grounding methods such as DAAM, OVAM, Seg4Diff, and ConceptAttention. These results show that the proposed anchor-based graph propagation improves localization quality not only in confusable multi-concept scenes, but also under the standard foreground segmentation protocol.

\begin{table}[t]
\centering

\begin{minipage}[t]{0.51\linewidth}
\renewcommand{\arraystretch}{0.9}
\centering
\captionof{table}{Results on the Multi-Concept Confusion Dataset.
We report $\mathrm{mIoU}_{fg}$, $\mathrm{mAP}_{fg}$, and $\mathrm{Acc}_{fg}$, averaged over image-concept pairs. Higher is better except for NAR.}
\vspace{0.1em}
\label{tab:multi_concept_results}
\footnotesize
\resizebox{\linewidth}{!}{
\begin{tabular}{
  p{3.05cm}|
  >{\centering\arraybackslash}p{0.9cm}
  >{\centering\arraybackslash}p{0.9cm}
  >{\centering\arraybackslash}p{0.7cm}
  >{\centering\arraybackslash}p{0.8cm}
}
\toprule
\textbf{Method}
& $\mathrm{mIoU}_{\mathrm{fg}}\uparrow$
& $\mathrm{mAP}_{\mathrm{fg}}\uparrow$
& $\mathrm{NAR}\downarrow$
& $\mathrm{Acc}_{\mathrm{fg}}\uparrow$\\
\midrule
CLIPasRNN              & 25.82 & 74.89 & 47.20 & 46.81 \\
TextSpan               & 25.47 & 59.53 & 45.99 & 52.38 \\
OVAM                   & 26.28 & 61.76 & 49.58 & 45.76 \\
DAAM SD2               & 27.10 & 71.03 & 44.08 & 47.02 \\
DAAM SDXL              & 24.46 & 64.12 & 50.66 & 45.31 \\
Cross Attention SD3    & 33.93 & 83.70 & 32.49 & 54.62 \\
Cross Attention SD3.5  & 28.37 & 88.06 & 28.45 & 65.97 \\
Seg4Diff SD3           & 44.56 & 82.06 & 22.47 & 65.88 \\
Seg4Diff SD3.5         & 46.57 & 84.22 & 19.08 & 69.19 \\
ConceptAttention SD3   & 38.24 & 80.63 & 32.95 & 55.26 \\
ConceptAttention SD3.5 & 39.97 & {88.73} & 26.36 & 59.28 \\
\midrule
\textbf{Ours SD3}        & {64.13} & {87.27} & {16.86} & {86.24} \\
\textbf{Ours SD3.5}      & \textbf{68.85}& \textbf{89.38} & \textbf{12.21} & \textbf{89.97} \\
\bottomrule
\end{tabular}
}
\end{minipage}
\hfill
\begin{minipage}[t]{0.47\linewidth}
\renewcommand{\arraystretch}{0.88}
\centering
\captionof{table}{Component ablation on the Multi-Concept Confusion Dataset. We evaluate the contribution of each component under the same inference setting. All variants are evaluated on the full dataset using the same metrics as \Cref{tab:multi_concept_results}.}
\label{tab:ablation_components}
\footnotesize
\resizebox{\linewidth}{!}{
\begin{tabular}{
  p{0.8cm}|
  >{\arraybackslash}p{2.0cm}|
  >{\centering\arraybackslash}p{0.9cm}
  >{\centering\arraybackslash}p{0.9cm}
  >{\centering\arraybackslash}p{0.8cm}
  >{\centering\arraybackslash}p{0.88cm}
}
\toprule
\textbf{Model} & {Settings} & $\mathrm{mIoU}_{\mathrm{fg}}\uparrow$
& $\mathrm{mAP}_{\mathrm{fg}}\uparrow$
& $\mathrm{NAR}\downarrow$
& $\mathrm{Acc}_{\mathrm{fg}}\uparrow$\\
\midrule
\multirow{6}{*}{SD3}
& ${A}_{C\to I}$ only & 34.61 & 71.66 & 35.51 & 54.91 \\
& w/o ${A}_{I\to I}$ gate & 7.63 & 9.06 & 55.11 & 50.31 \\
& w/o ${W}_{\mathrm{cos}}$ & 63.01 & 86.87 & {16.72} & 86.21 \\
& L9 only & 55.32 & 80.47 & 13.78 & 84.32 \\
& L18 only & 42.28 & 82.06 & 25.23 & 70.92 \\
& \textbf{Ours full} & {64.13} & {87.27} & {16.86} & {86.24} \\
\midrule
\multirow{6}{*}{SD3.5}
& ${A}_{C\to I}$ only & 29.91 & 80.11 & 37.82 & 51.82 \\
& w/o ${A}_{I\to I}$ gate & 11.97 & 14.42 & 55.05 & 49.78 \\
& w/o ${W}_{\mathrm{cos}}$ & 67.63 & 88.97 & {12.21} & 89.93 \\
& L23 only & 57.00 & 79.91 & 10.75 & 88.82 \\
& L31 only & 53.44 & 86.36 & 17.59 & 79.19 \\
& \textbf{Ours full} & {68.85}& {89.38} & {12.20} & {89.97}\\
\bottomrule
\end{tabular}
}
\end{minipage}
\vspace{-0.8em}
\end{table}

\vspace{-0.5em}
\subsection{Ablation studies}
\vspace{-0.5em}
\label{sec:exp:ablation}

We conduct ablation studies to analyze the main components, propagation steps, and layer selection on the Multi-Concept Confusion Dataset, and the effect of auxiliary background concepts on ImageNet-Segmentation.

\begin{figure}[t]
\centering

\begin{minipage}[t]{0.48\linewidth}
\vspace{0pt}
\centering
\includegraphics[width=\linewidth]{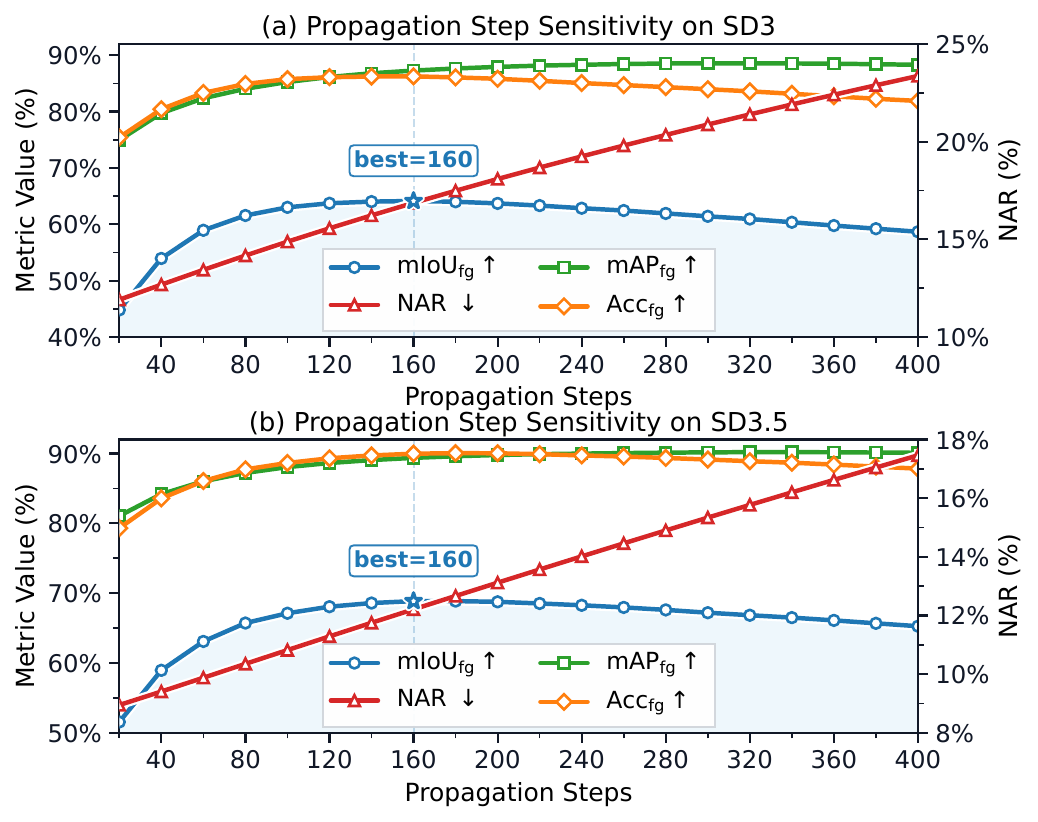}
\captionof{figure}{Propagation step sensitivity on SD3 and SD3.5.}
\label{fig:diffusion_step_sensitivity}
\end{minipage}%
\hfill%
\begin{minipage}[t]{0.48\linewidth}
\vspace{0pt}
\centering
\captionof{table}{
Single-concept grounding on ImageNet-Segmentation using SD3 and SD3.5. We compare ConceptAttention and AnchorDiff with and without auxiliary background concepts. AnchorDiff works directly with a single target concept, and auxiliary background concepts have only marginal impact across backbones.
}
\label{tab:single_concept}
\vspace{-2pt}

\footnotesize
\renewcommand{\arraystretch}{0.9}
\resizebox{\linewidth}{!}{%
\begin{tabular}
{p{0.8cm}|p{4.1cm}|
  >{\centering\arraybackslash}p{0.85cm}
  >{\centering\arraybackslash}p{0.85cm}}
\toprule
\textbf{Model} & {Settings} & mIoU$\uparrow$ & Acc$\uparrow$\\
\midrule
\multirow{5}{*}{SD3}
&ConceptAttention (single) & 54.71 & 70.72 \\
&ConceptAttention (w/o softmax) & 58.87 & 74.11 \\
&ConceptAttention (with BG) & 68.75 & 81.48 \\
&\textbf{Ours (single)} & 75.09 & 83.76 \\
&\textbf{Ours (with BG)} & 74.88 & 83.73 \\
\midrule
\multirow{5}{*}{SD3.5}
&ConceptAttention (single) & 55.24 & 71.24 \\
&ConceptAttention (w/o softmax) & 58.68 & 73.96 \\
&ConceptAttention (with BG) & 72.04 & 83.75 \\
&\textbf{Ours (single)} & 77.29 & 85.46 \\
&\textbf{Ours (with BG)} & 77.54 & 85.63 \\
\bottomrule
\end{tabular}%
}

\end{minipage}

\vspace{-0.5em}
\end{figure}

\vspace{-0.5em}
\paragraph{Main components.}
\Cref{tab:ablation_components} evaluates the main components of our pipeline. The variant ${A}_{C\to I}$-only directly uses the concept-to-image attention map without graph propagation. Its substantially lower $\mathrm{mIoU}_{fg}$ indicates that the raw concept-to-image attention map alone is insufficient for separating confusable objects.
Removing the ${A}_{I\to I}$ attention gate leads to the largest degradation
for both backbones. This indicates that output-space similarity alone is not target-specific and can propagate activation to visually or semantically similar objects. In contrast, removing ${W}_{\mathrm{cos}}$ remains competitive but underperforms the full model in $\mathrm{mIoU}_{fg}$ and $\mathrm{mAP}_{fg}$, suggesting that the row-wise attention gate suppresses cross-object leakage,
while output-space affinity improves within-object propagation.

\vspace{-0.5em}
\paragraph{Propagation steps.}
We study the effect of the number of graph propagation steps in
\Cref{fig:diffusion_step_sensitivity}. With all other parameters fixed,
$\mathrm{mIoU}_{\mathrm{fg}}$ and $\mathrm{Acc}_{\mathrm{fg}}$ rise rapidly at early steps and saturate around 160 steps. After that, both metrics gradually decrease, while NAR continues to increase, indicating that excessive propagation can spread activation beyond the target object through weak residual connections. $\mathrm{mAP}_{\mathrm{fg}}$ also rises quickly at early steps and then remains relatively stable. These results suggest that a moderate propagation length is sufficient for object coverage, whereas overly long propagation reduces object specificity.

\begin{figure}[t]
\centering
\includegraphics[width=\linewidth]{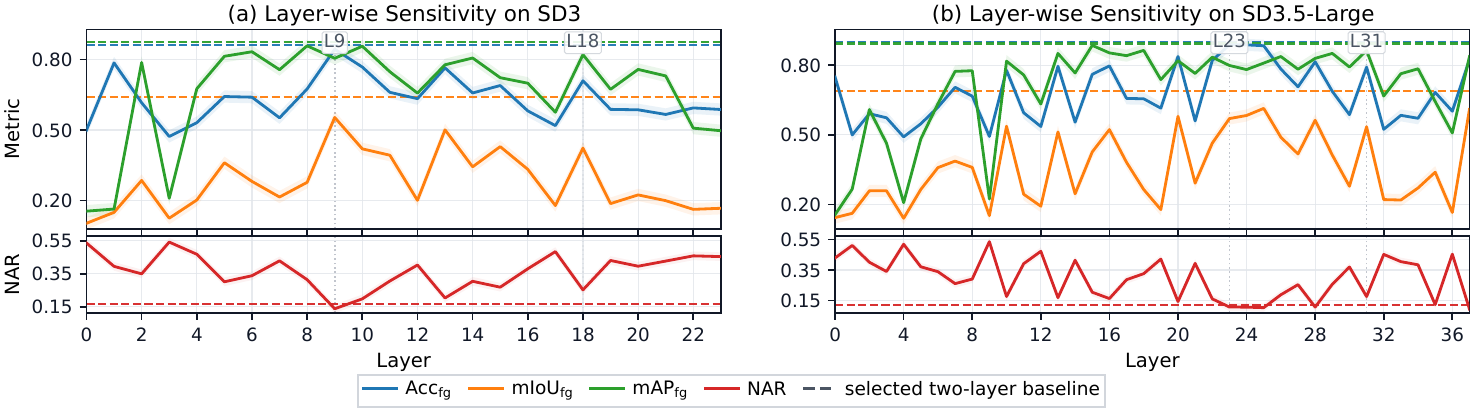}
\caption{
Layer-wise sensitivity analysis on the Multi-Concept Confusion Dataset.
We evaluate single-layer image-to-image attention features across the transformer
depth for SD3 and SD3.5-Large.
Dashed lines indicate the selected two-layer configurations used in the full
method.
}
\label{fig:layer_sensitivity}
\vspace{-1em}
\end{figure}

\vspace{-0.5em}
\paragraph{Layer selection.}
We analyze the effect of image-to-image layer selection in
\Cref{tab:ablation_components} and \Cref{fig:layer_sensitivity}. The single-layer results show that different layers provide complementary cues. L9 is stronger than L18 for SD3, while L23 and L31 show different trade-offs for SD3.5. Their two-layer fusions achieve the best $\mathrm{mIoU}_{\mathrm{fg}}$ among two-layer configurations. We additionally evaluate all two-layer combinations among the top-performing single layers, as well as randomly sampled three-layer and four-layer combinations, with the full sweep on the Multi-Concept Confusion Dataset reported in \Cref{sec:layers_analysis} of the appendix. The selected two-layer pairs, L9+L18 for SD3 and L23+L31 for SD3.5, consistently outperform other two-layer configurations across all three benchmarks. While a few multi-layer variants achieve marginally higher scores, the gains are small and we adopt the simpler two-layer configuration in the main experiments. This suggests that layer fusion benefits from complementary structural cues across depths rather than simply selecting the strongest individual layers.

\vspace{-0.5em}
\paragraph{Single-concept analysis.}
\Cref{tab:single_concept} evaluates the effect of auxiliary background concepts. ConceptAttention degrades with a single concept and benefits substantially from auxiliary background concepts, while AnchorDiff remains strong without them. Unlike methods that require background concepts to create spatial contrast, AnchorDiff uses the target response only to select a sparse anchor and obtains the mask through structural propagation. The marginal changes from adding background concepts indicate that AnchorDiff does not rely on auxiliary background concepts.

\vspace{-0.5em}
\section{Conclusion}
\vspace{-0.5em}
\label{sec:conclusion}

We presented {AnchorDiff}, a training-free semantic grounding method for MM-DiT models. AnchorDiff selects a high-confidence target-concept anchor and propagates a one-hot seed over a hybrid graph built from image-to-image self-attention and output-space features, decoupling semantic localization from structural refinement. Experiments show that AnchorDiff achieves strong grounding performance on ImageNet-Segmentation and PascalVOC, while substantially reducing concept leakage on our Multi-Concept Confusion Dataset. One limitation is that AnchorDiff can inherit errors from the initial semantic anchor when concept-to-image attention is unreliable, suggesting that more robust semantic initialization remains important for training-free grounding. Additionally, since AnchorDiff selects a single anchor per concept, it may fail to cover all instances when multiple same-category instances are spatially distant, as a single seed cannot reach disconnected regions. Extending it to multi-anchor selection is a natural direction for future work.

\bibliographystyle{unsrtnat}
\bibliography{references}

\begin{thebibliography}{30}
\providecommand{\natexlab}[1]{#1}
\providecommand{\url}[1]{\texttt{#1}}
\expandafter\ifx\csname urlstyle\endcsname\relax
  \providecommand{\doi}[1]{doi: #1}\else
  \providecommand{\doi}{doi: \begingroup \urlstyle{rm}\Url}\fi

\bibitem[Ho et~al.(2020)Ho, Jain, and Abbeel]{ho2020denoising}
Jonathan Ho, Ajay Jain, and Pieter Abbeel.
\newblock Denoising diffusion probabilistic models.
\newblock \emph{Advances in neural information processing systems}, 33:\penalty0 6840--6851, 2020.

\bibitem[Rombach et~al.(2022)Rombach, Blattmann, Lorenz, Esser, and Ommer]{rombach2022high}
Robin Rombach, Andreas Blattmann, Dominik Lorenz, Patrick Esser, and Bj{\"o}rn Ommer.
\newblock High-resolution image synthesis with latent diffusion models.
\newblock In \emph{Proceedings of the IEEE/CVF conference on computer vision and pattern recognition}, pages 10684--10695, 2022.

\bibitem[Tang et~al.(2022)Tang, Liu, Pandey, Jiang, Yang, Kumar, Stenetorp, Lin, and Ture]{tang2022daam}
Raphael Tang, Linqing Liu, Akshat Pandey, Zhiying Jiang, Gefei Yang, Karun Kumar, Pontus Stenetorp, Jimmy Lin, and Ferhan Ture.
\newblock What the daam: Interpreting stable diffusion using cross attention.
\newblock \emph{arXiv preprint arXiv:2210.04885}, 2022.

\bibitem[Hertz et~al.(2022)Hertz, Mokady, Tenenbaum, Aberman, Pritch, and Cohen-Or]{hertz2022prompt}
Amir Hertz, Ron Mokady, Jay Tenenbaum, Kfir Aberman, Yael Pritch, and Daniel Cohen-Or.
\newblock Prompt-to-prompt image editing with cross attention control.
\newblock \emph{arXiv preprint arXiv:2208.01626}, 2022.

\bibitem[Peebles and Xie(2023)]{peebles2023scalable}
William Peebles and Saining Xie.
\newblock Scalable diffusion models with transformers.
\newblock In \emph{Proceedings of the IEEE/CVF international conference on computer vision}, pages 4195--4205, 2023.

\bibitem[Helbling et~al.(2025)Helbling, Meral, Hoover, Yanardag, and Chau]{helbling2025conceptattention}
Alec Helbling, Tuna Meral, Benjamin Hoover, Pinar Yanardag, and Polo Chau.
\newblock Conceptattention: Diffusion transformers learn highly interpretable features.
\newblock In \emph{International Conference on Machine Learning}, 2025.

\bibitem[Kim et~al.(2025)Kim, Shin, Hong, Yoon, Arnab, Seo, Hong, and Kim]{kim2025seg4diff}
Chaehyun Kim, Heeseong Shin, Eunbeen Hong, Heeji Yoon, Anurag Arnab, Paul~Hongsuck Seo, Sunghwan Hong, and Seungryong Kim.
\newblock Seg4diff: Unveiling open-vocabulary semantic segmentation in text-to-image diffusion transformers.
\newblock In \emph{The Thirty-ninth Annual Conference on Neural Information Processing Systems}, 2025.

\bibitem[Esser et~al.(2024)Esser, Kulal, Blattmann, Entezari, M{\"u}ller, Saini, Levi, Lorenz, Sauer, Boesel, et~al.]{esser2024scaling}
Patrick Esser, Sumith Kulal, Andreas Blattmann, Rahim Entezari, Jonas M{\"u}ller, Harry Saini, Yam Levi, Dominik Lorenz, Axel Sauer, Frederic Boesel, et~al.
\newblock Scaling rectified flow transformers for high-resolution image synthesis.
\newblock \emph{arXiv e-prints}, pages arXiv--2403, 2024.

\bibitem[Radford et~al.(2021)Radford, Kim, Hallacy, Ramesh, Goh, Agarwal, Sastry, Askell, Mishkin, Clark, et~al.]{radford2021learning}
Alec Radford, Jong~Wook Kim, Chris Hallacy, Aditya Ramesh, Gabriel Goh, Sandhini Agarwal, Girish Sastry, Amanda Askell, Pamela Mishkin, Jack Clark, et~al.
\newblock Learning transferable visual models from natural language supervision.
\newblock In \emph{International conference on machine learning}, pages 8748--8763. PmLR, 2021.

\bibitem[Chefer et~al.(2021)Chefer, Gur, and Wolf]{chefer2021transformer}
Hila Chefer, Shir Gur, and Lior Wolf.
\newblock Transformer interpretability beyond attention visualization.
\newblock In \emph{Proceedings of the IEEE/CVF conference on computer vision and pattern recognition}, pages 782--791, 2021.

\bibitem[Selvaraju et~al.(2017)Selvaraju, Cogswell, Das, Vedantam, Parikh, and Batra]{selvaraju2017grad}
Ramprasaath~R Selvaraju, Michael Cogswell, Abhishek Das, Ramakrishna Vedantam, Devi Parikh, and Dhruv Batra.
\newblock Grad-cam: Visual explanations from deep networks via gradient-based localization.
\newblock In \emph{Proceedings of the IEEE international conference on computer vision}, pages 618--626, 2017.

\bibitem[Gandelsman et~al.(2023)Gandelsman, Efros, and Steinhardt]{gandelsman2023interpreting}
Yossi Gandelsman, Alexei~A Efros, and Jacob Steinhardt.
\newblock Interpreting clip's image representation via text-based decomposition.
\newblock \emph{arXiv preprint arXiv:2310.05916}, 2023.

\bibitem[Sun et~al.(2024)Sun, Li, Torr, Gu, and Li]{sun2024clip}
Shuyang Sun, Runjia Li, Philip Torr, Xiuye Gu, and Siyang Li.
\newblock Clip as rnn: Segment countless visual concepts without training endeavor.
\newblock In \emph{Proceedings of the IEEE/CVF Conference on Computer Vision and Pattern Recognition}, pages 13171--13182, 2024.

\bibitem[Vaswani et~al.(2017)Vaswani, Shazeer, Parmar, Uszkoreit, Jones, Gomez, Kaiser, and Polosukhin]{2017Attention}
Ashish Vaswani, Noam Shazeer, Niki Parmar, Jakob Uszkoreit, Llion Jones, Aidan~N Gomez, Lukasz Kaiser, and Illia Polosukhin.
\newblock Attention is all you need.
\newblock \emph{arXiv}, 2017.

\bibitem[Wang et~al.(2025)Wang, Li, Zhang, Xu, Zhou, Yu, Sheng, and Xu]{wang2025diffusion}
Jinglong Wang, Xiawei Li, Jing Zhang, Qingyuan Xu, Qin Zhou, Qian Yu, Lu~Sheng, and Dong Xu.
\newblock Diffusion model is secretly a training-free open vocabulary semantic segmenter.
\newblock \emph{IEEE Transactions on Image Processing}, 2025.

\bibitem[Marcos-Manch{\'o}n et~al.(2024)Marcos-Manch{\'o}n, Alcover-Couso, SanMiguel, and Mart{\'\i}nez]{marcos2024open}
Pablo Marcos-Manch{\'o}n, Roberto Alcover-Couso, Juan~C SanMiguel, and Jose~M Mart{\'\i}nez.
\newblock Open-vocabulary attention maps with token optimization for semantic segmentation in diffusion models.
\newblock In \emph{Proceedings of the IEEE/CVF conference on computer vision and pattern recognition}, pages 9242--9252, 2024.

\bibitem[Podell et~al.()Podell, English, Lacey, Blattmann, Dockhorn, M{\"u}ller, Penna, and Rombach]{podellsdxl}
Dustin Podell, Zion English, Kyle Lacey, Andreas Blattmann, Tim Dockhorn, Jonas M{\"u}ller, Joe Penna, and Robin Rombach.
\newblock Sdxl: Improving latent diffusion models for high-resolution image synthesis.
\newblock In \emph{The Twelfth International Conference on Learning Representations}.

\bibitem[Guillaumin et~al.(2014)Guillaumin, K{\"u}ttel, and Ferrari]{guillaumin2014imagenet}
Matthieu Guillaumin, Daniel K{\"u}ttel, and Vittorio Ferrari.
\newblock Imagenet auto-annotation with segmentation propagation.
\newblock \emph{International Journal of Computer Vision}, 110\penalty0 (3):\penalty0 328--348, 2014.

\bibitem[Everingham et~al.(2015)Everingham, Eslami, Van~Gool, Williams, Winn, and Zisserman]{2015The}
Mark Everingham, S.~M.~Ali Eslami, Luc Van~Gool, Christopher K.~I. Williams, John Winn, and Andrew Zisserman.
\newblock The pascal visual object classes challenge: A retrospective.
\newblock \emph{International Journal of Computer Vision}, 111\penalty0 (1):\penalty0 98--136, 2015.

\bibitem[Gao et~al.(2025)Gao, Gong, Guo, Hou, Lai, Li, Li, Lian, Liao, Liu, et~al.]{gao2025seedream}
Yu~Gao, Lixue Gong, Qiushan Guo, Xiaoxia Hou, Zhichao Lai, Fanshi Li, Liang Li, Xiaochen Lian, Chao Liao, Liyang Liu, et~al.
\newblock Seedream 3.0 technical report.
\newblock \emph{arXiv preprint arXiv:2504.11346}, 2025.

\bibitem[Seedream et~al.(2025)Seedream, Chen, Gao, Gong, Guo, Guo, Guo, Hou, Huang, Huang, et~al.]{seedream2025seedream}
Team Seedream, Yunpeng Chen, Yu~Gao, Lixue Gong, Meng Guo, Qiushan Guo, Zhiyao Guo, Xiaoxia Hou, Weilin Huang, Yixuan Huang, et~al.
\newblock Seedream 4.0: Toward next-generation multimodal image generation.
\newblock \emph{arXiv preprint arXiv:2509.20427}, 2025.

\bibitem[Bai et~al.(2025)Bai, Cai, Chen, Chen, Chen, Cheng, Deng, Ding, Gao, Ge, et~al.]{bai2025qwen3}
Shuai Bai, Yuxuan Cai, Ruizhe Chen, Keqin Chen, Xionghui Chen, Zesen Cheng, Lianghao Deng, Wei Ding, Chang Gao, Chunjiang Ge, et~al.
\newblock Qwen3-vl technical report.
\newblock \emph{arXiv preprint arXiv:2511.21631}, 2025.

\bibitem[Ravi et~al.()Ravi, Gabeur, Hu, Hu, Ryali, Ma, Khedr, R{\"a}dle, Rolland, Gustafson, et~al.]{ravisam}
Nikhila Ravi, Valentin Gabeur, Yuan-Ting Hu, Ronghang Hu, Chaitanya Ryali, Tengyu Ma, Haitham Khedr, Roman R{\"a}dle, Chloe Rolland, Laura Gustafson, et~al.
\newblock Sam 2: Segment anything in images and videos.
\newblock In \emph{The Thirteenth International Conference on Learning Representations}.

\bibitem[Binder et~al.(2016)Binder, Montavon, Lapuschkin, M{\"u}ller, and Samek]{binder2016layer}
Alexander Binder, Gr{\'e}goire Montavon, Sebastian Lapuschkin, Klaus-Robert M{\"u}ller, and Wojciech Samek.
\newblock Layer-wise relevance propagation for neural networks with local renormalization layers.
\newblock In \emph{International conference on artificial neural networks}, pages 63--71. Springer, 2016.

\bibitem[Abnar and Zuidema(2005)]{abnar2005quantifying}
Samira Abnar and Willem Zuidema.
\newblock Quantifying attention flow in transformers. arxiv 2020.
\newblock \emph{arXiv preprint arXiv:2005.00928}, 10, 2005.

\bibitem[Dosovitskiy et~al.(2020)Dosovitskiy, Beyer, Kolesnikov, Weissenborn, and Houlsby]{2020An}
Alexey Dosovitskiy, Lucas Beyer, Alexander Kolesnikov, Dirk Weissenborn, and Neil Houlsby.
\newblock An image is worth 16x16 words: Transformers for image recognition at scale.
\newblock 2020.

\bibitem[Caron et~al.(2021)Caron, Touvron, Misra, J{\'e}gou, Mairal, Bojanowski, and Joulin]{caron2021emerging}
Mathilde Caron, Hugo Touvron, Ishan Misra, Herv{\'e} J{\'e}gou, Julien Mairal, Piotr Bojanowski, and Armand Joulin.
\newblock Emerging properties in self-supervised vision transformers.
\newblock In \emph{Proceedings of the IEEE/CVF international conference on computer vision}, pages 9650--9660, 2021.

\bibitem[Oquab et~al.(2024)Oquab, Darcet, Moutakanni, Vo, Szafraniec, Khalidov, Fernandez, Haziza, Massa, El-Nouby, et~al.]{oquab2024dinov2}
Maxime Oquab, Timoth{\'e}e Darcet, Th{\'e}o Moutakanni, Huy Vo, Marc Szafraniec, Vasil Khalidov, Pierre Fernandez, Daniel Haziza, Francisco Massa, Alaaeldin El-Nouby, et~al.
\newblock Dinov2: Learning robust visual features without supervision.
\newblock \emph{Transactions on Machine Learning Research Journal}, 2024.

\bibitem[Darcet et~al.(2024)Darcet, Oquab, Mairal, and Bojanowski]{darcet2024vision}
Timoth{\'e}e Darcet, Maxime Oquab, Julien Mairal, and Piotr Bojanowski.
\newblock Vision transformers need registers.
\newblock In \emph{International Conference on Learning Representations (ICLR)}, 2024.

\bibitem[Zhou et~al.(2021)Zhou, Wei, Wang, Shen, Xie, Yuille, and Kong]{zhou2021ibot}
Jinghao Zhou, Chen Wei, Huiyu Wang, Wei Shen, Cihang Xie, Alan Yuille, and Tao Kong.
\newblock ibot: Image bert pre-training with online tokenizer.
\newblock \emph{arXiv preprint arXiv:2111.07832}, 2021.

\end{thebibliography}

\newpage
\appendix

\section{Technical appendices and supplementary material}

\subsection{Additional dataset samples}
\begin{figure}[H]
\vspace{-1.3em}
\centering
\includegraphics[
  width=1\textwidth,
  trim=10 20 20 10,
  clip
]{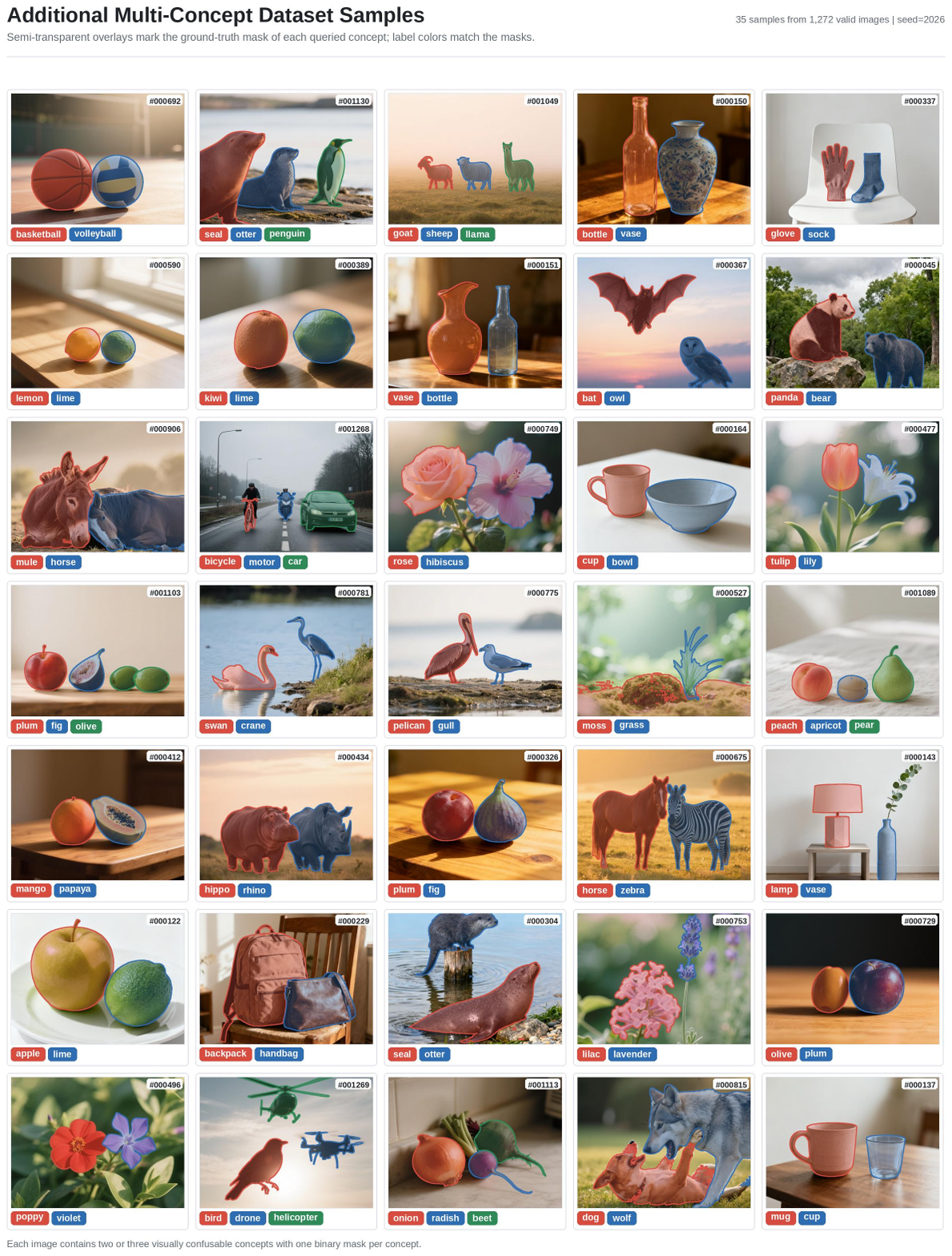}
\vspace{-1.5em}
\caption{
Additional samples from the Multi-Concept Confusion Dataset. Each image contains two or three visually confusable concepts with one binary mask per concept.
}
\label{fig:app_dataset_samples}
\end{figure}

\Cref{fig:app_dataset_samples} provides more examples from the Multi-Concept Confusion Dataset. Each image contains two or three visually confusable concepts with separate per-concept masks, shown as semi-transparent overlays. These samples highlight the fine-grained visual ambiguities used to evaluate concept leakage.

\clearpage

\subsection{Additional qualitative comparisons}
\vspace{-0.5em}
\label{app:qualitative}

\paragraph{Two-concept qualitative comparisons.}
We provide additional qualitative comparisons on two-concept confusion examples. Each example contains two visually similar concepts, and each concept is evaluated independently with both heatmap and thresholded-mask visualizations. AnchorDiff produces more target-confined responses and reduces activation leakage to visually similar non-target concepts.

\vspace{-0.5em}
\paragraph{Three-concept qualitative comparisons.}
We provide additional qualitative comparisons on three-concept confusion examples. These examples are more challenging because the queried concept must be separated from two visually or semantically similar non-target concepts. AnchorDiff remains more object-confined across categories, while other methods often activate multiple similar objects.

\begin{figure}[H]
\vspace{-1em}
\centering
\includegraphics[
  page=1,
  width=0.95\textwidth,
  trim=200 10 200 10,
  clip
]{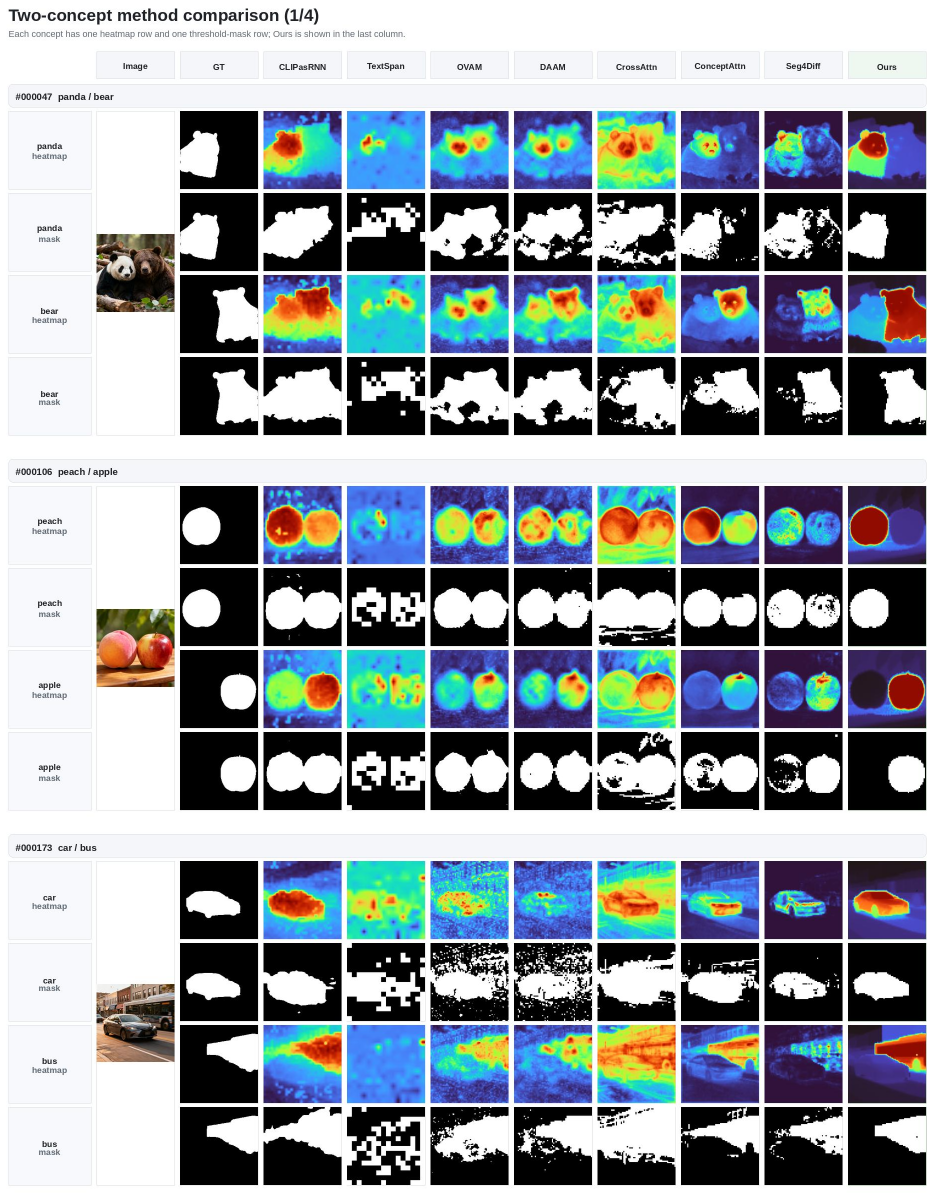}
\caption{
Additional qualitative comparisons on two-concept samples from the Multi-Concept Confusion Dataset. Page 1 of 4.
}
\label{fig:app_two_concept_1}
\end{figure}
\clearpage

\clearpage
\begin{figure*}[p]
\centering
\includegraphics[
  page=2,
  width=\textwidth,
  keepaspectratio,
  trim=200 10 200 10,
  clip
]{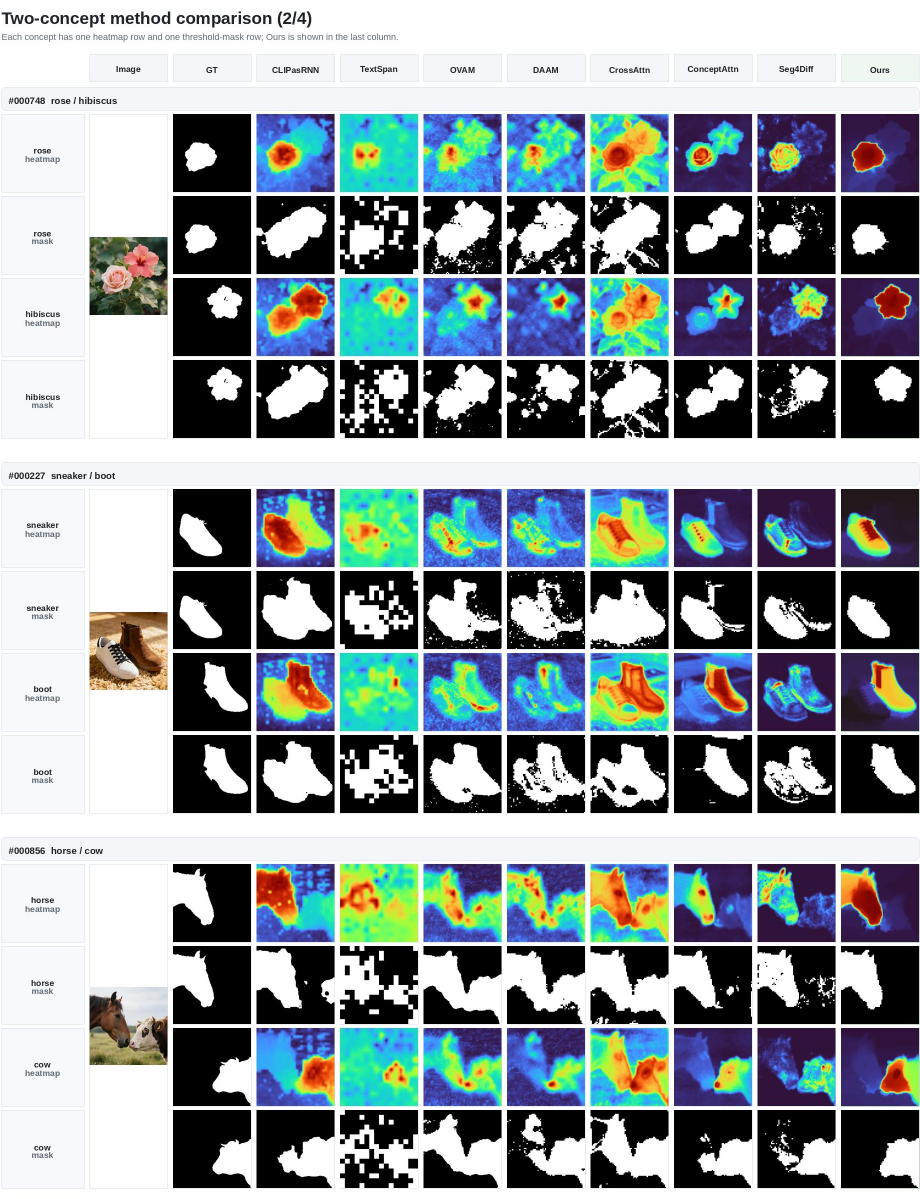}
\caption{
Additional qualitative comparisons on two-concept samples from the Multi-Concept Confusion Dataset. Page 2 of 4.
}
\label{fig:app_two_concept_2}
\end{figure*}

\begin{figure*}[p]
\centering
\includegraphics[
  page=3,
  width=\textwidth,
  height=0.90\textheight,
  keepaspectratio,
  trim=200 10 200 10,
  clip
]{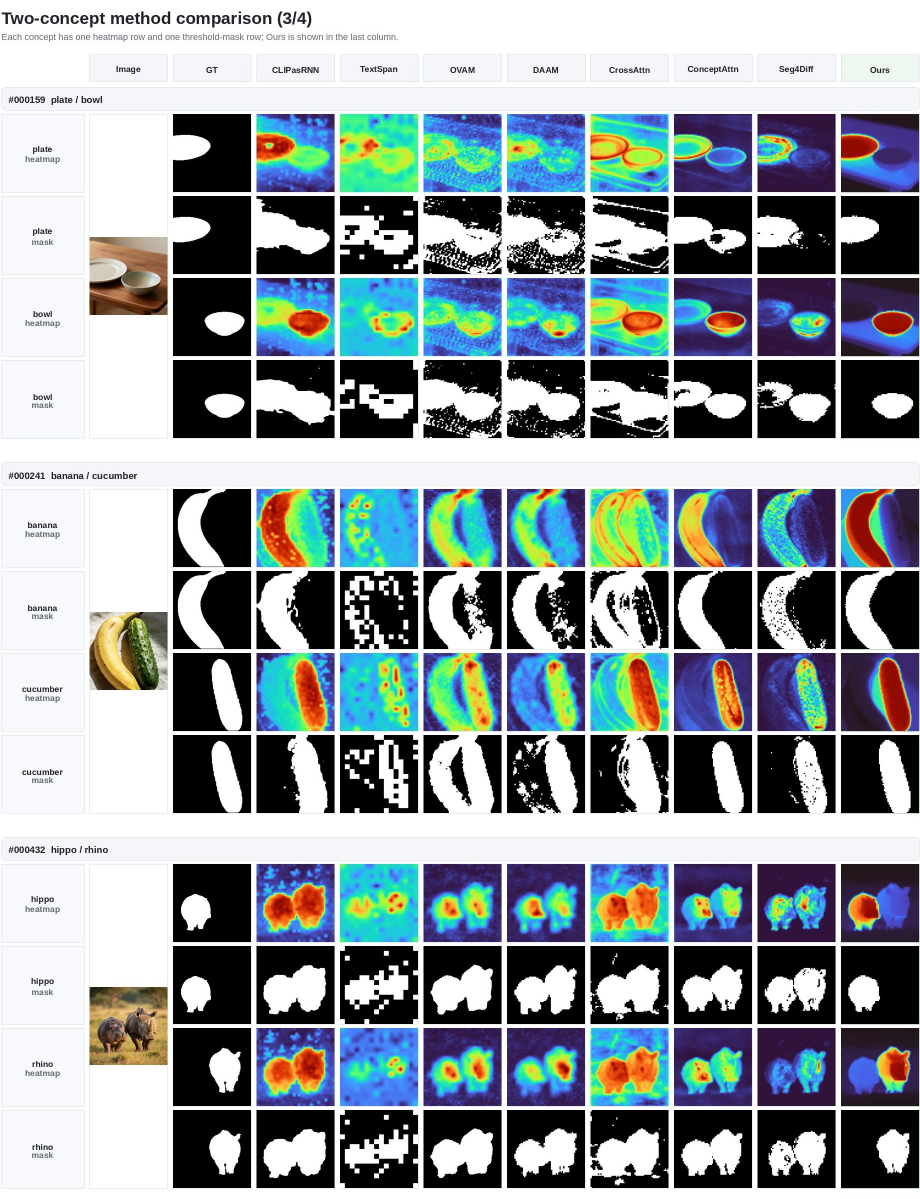}
\caption{
Additional qualitative comparisons on two-concept samples from the Multi-Concept Confusion Dataset. Page 3 of 4.
}
\label{fig:app_two_concept_3}
\end{figure*}
\clearpage

\clearpage
\begin{figure*}[p]
\centering
\includegraphics[
  page=4,
  width=\textwidth,
  height=0.90\textheight,
  keepaspectratio,
  trim=200 10 200 10,
  clip
]{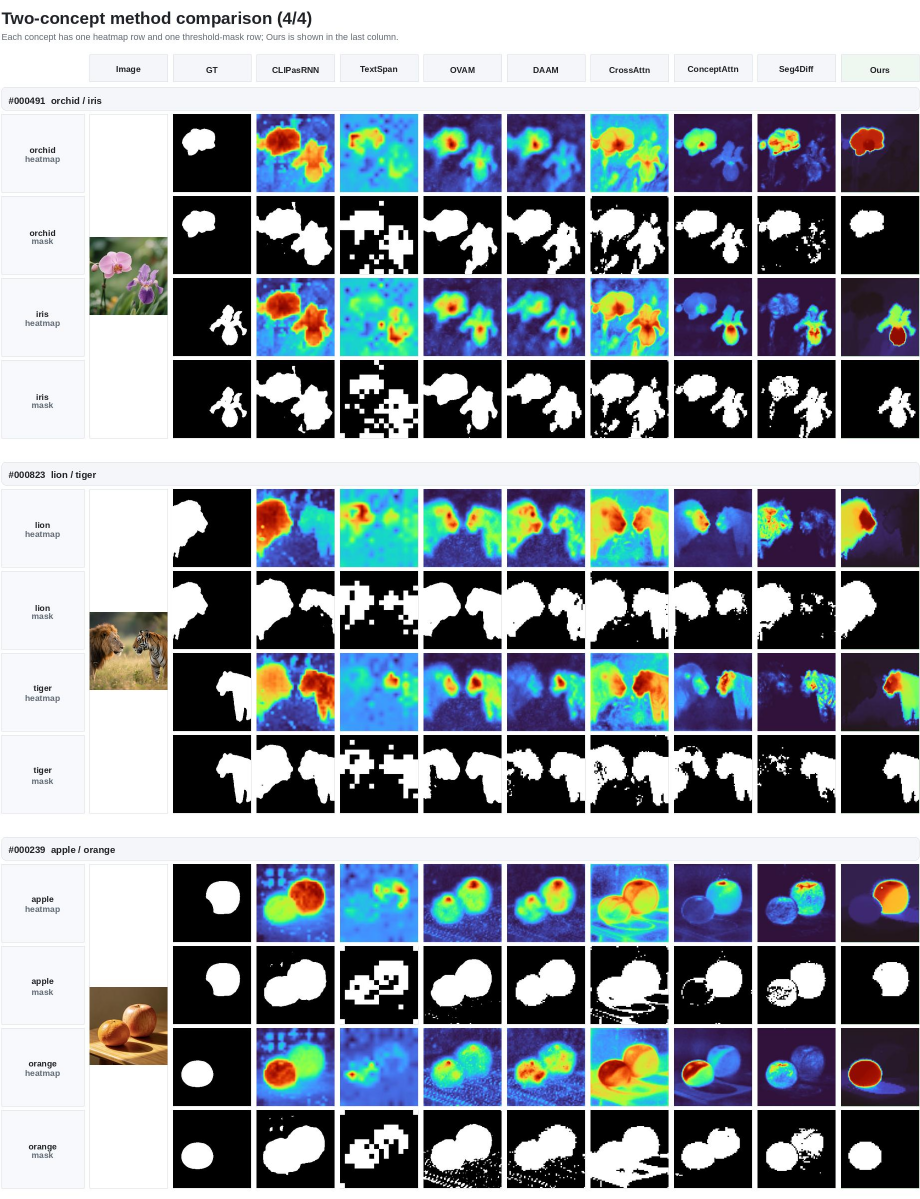}
\caption{
Additional qualitative comparisons on two-concept samples from the Multi-Concept Confusion Dataset. Page 4 of 4.
}
\label{fig:app_two_concept_4}
\end{figure*}
\clearpage

\begin{figure*}[p]
\centering
\includegraphics[
  page=1,
  width=\textwidth,
  height=0.90\textheight,
  keepaspectratio,
  trim=190 10 190 10,
  clip
]{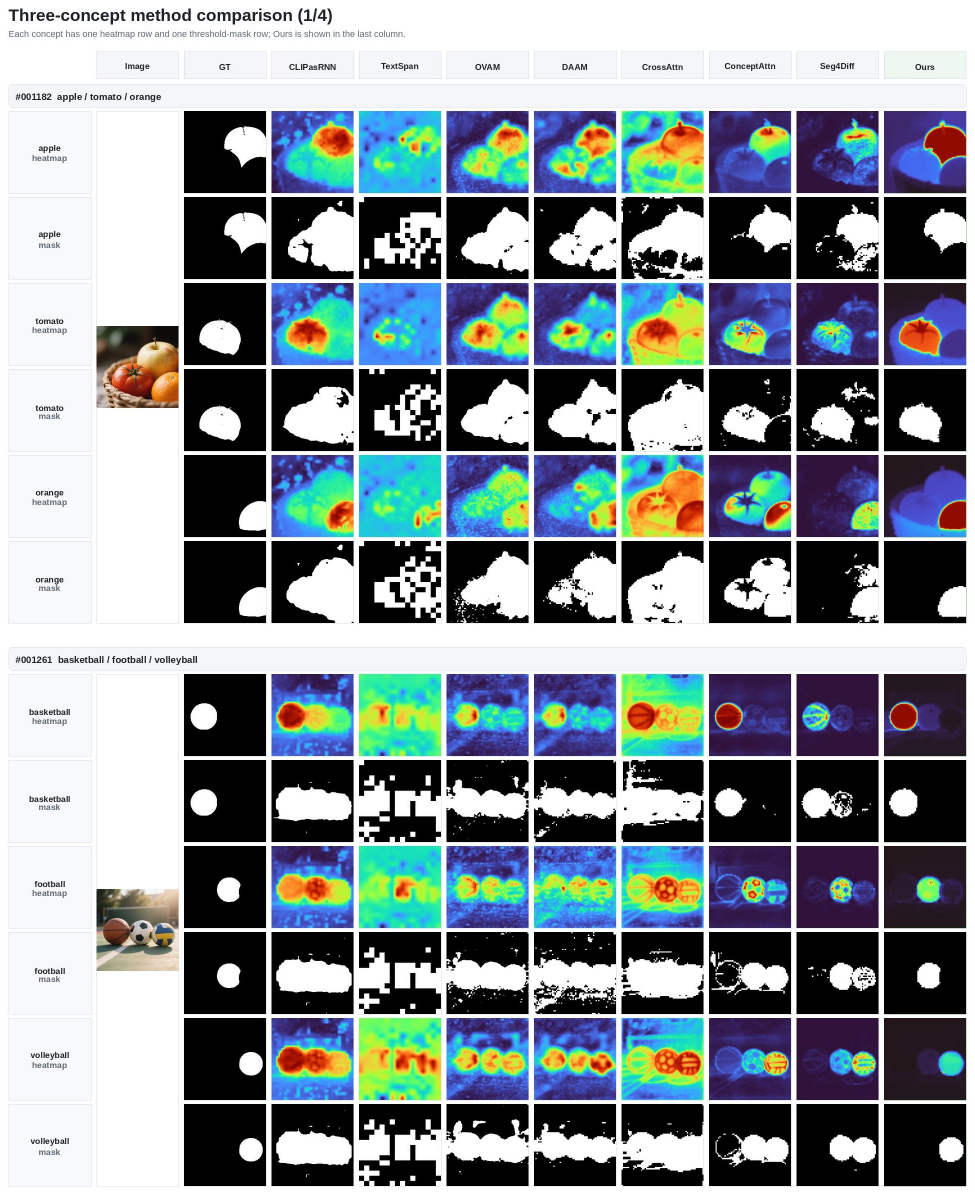}
\caption{
Additional qualitative comparisons on three-concept samples from the Multi-Concept Confusion Dataset. Page 1 of 4.
}
\label{fig:app_three_concept_1}
\end{figure*}
\clearpage

\clearpage
\begin{figure*}[p]
\centering
\includegraphics[
  page=2,
  width=\textwidth,
  height=0.90\textheight,
  keepaspectratio,
  trim=190 10 190 10,
  clip
]{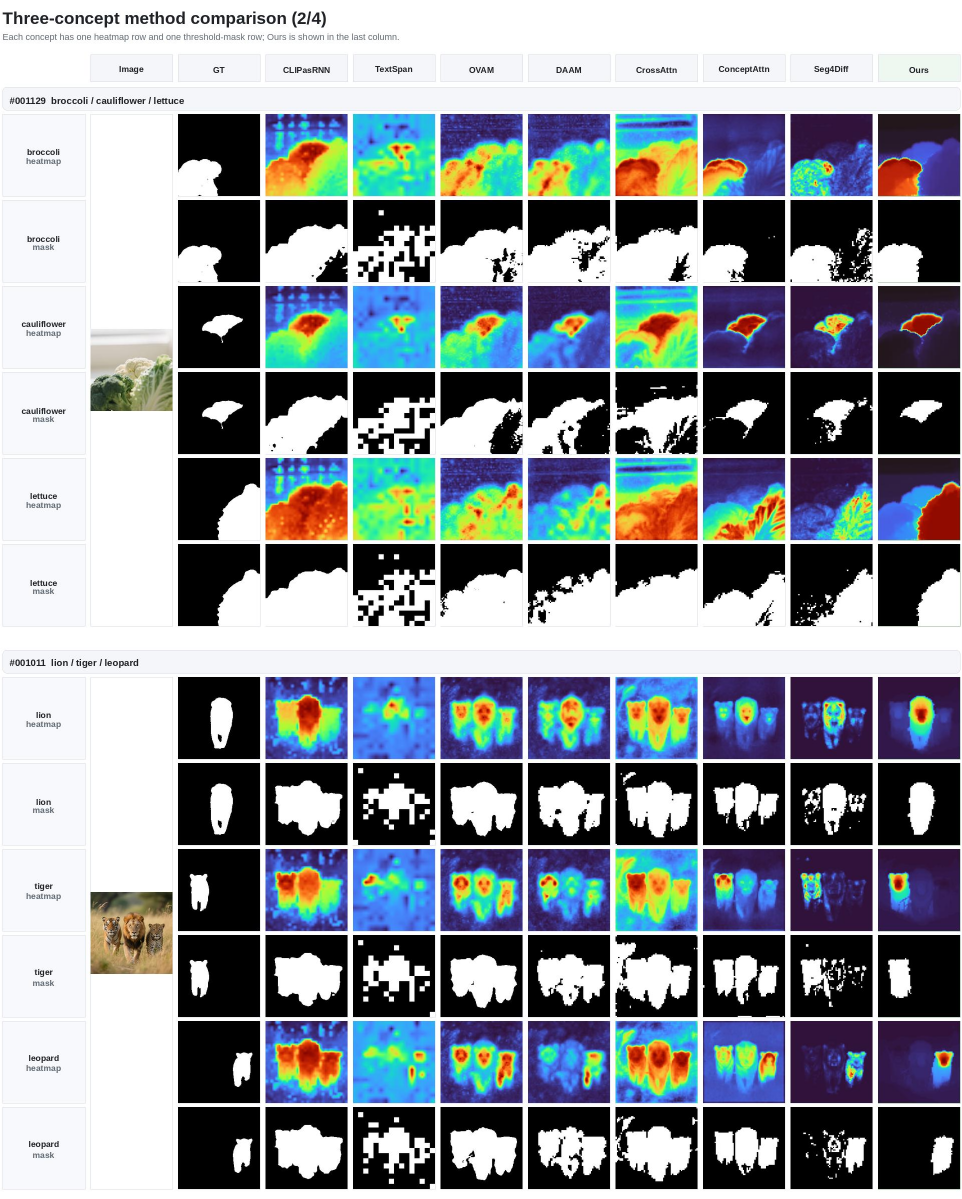}
\caption{
Additional qualitative comparisons on three-concept samples from the Multi-Concept Confusion Dataset. Page 2 of 4.
}
\label{fig:app_three_concept_2}
\end{figure*}
\clearpage

\clearpage
\begin{figure*}[p]
\centering
\includegraphics[
  page=3,
  width=\textwidth,
  height=0.90\textheight,
  keepaspectratio,
  trim=190 10 190 10,
  clip
]{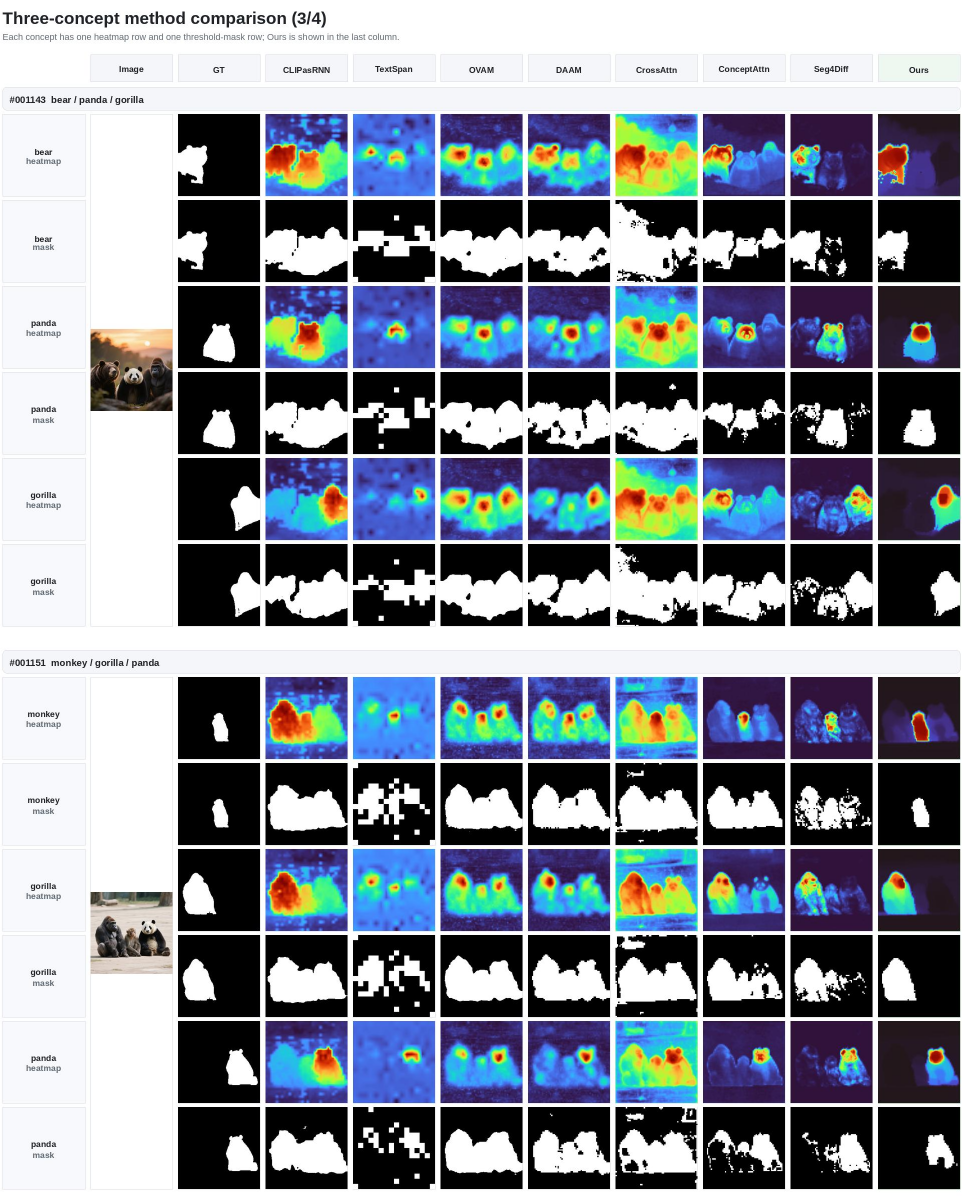}
\caption{
Additional qualitative comparisons on three-concept samples from the Multi-Concept Confusion Dataset. Page 3 of 4.
}
\label{fig:app_three_concept_3}
\end{figure*}
\clearpage

\clearpage
\begin{figure*}[p]
\centering
\includegraphics[
  page=4,
  width=\textwidth,
  height=0.90\textheight,
  keepaspectratio,
  trim=190 10 190 10,
  clip
]{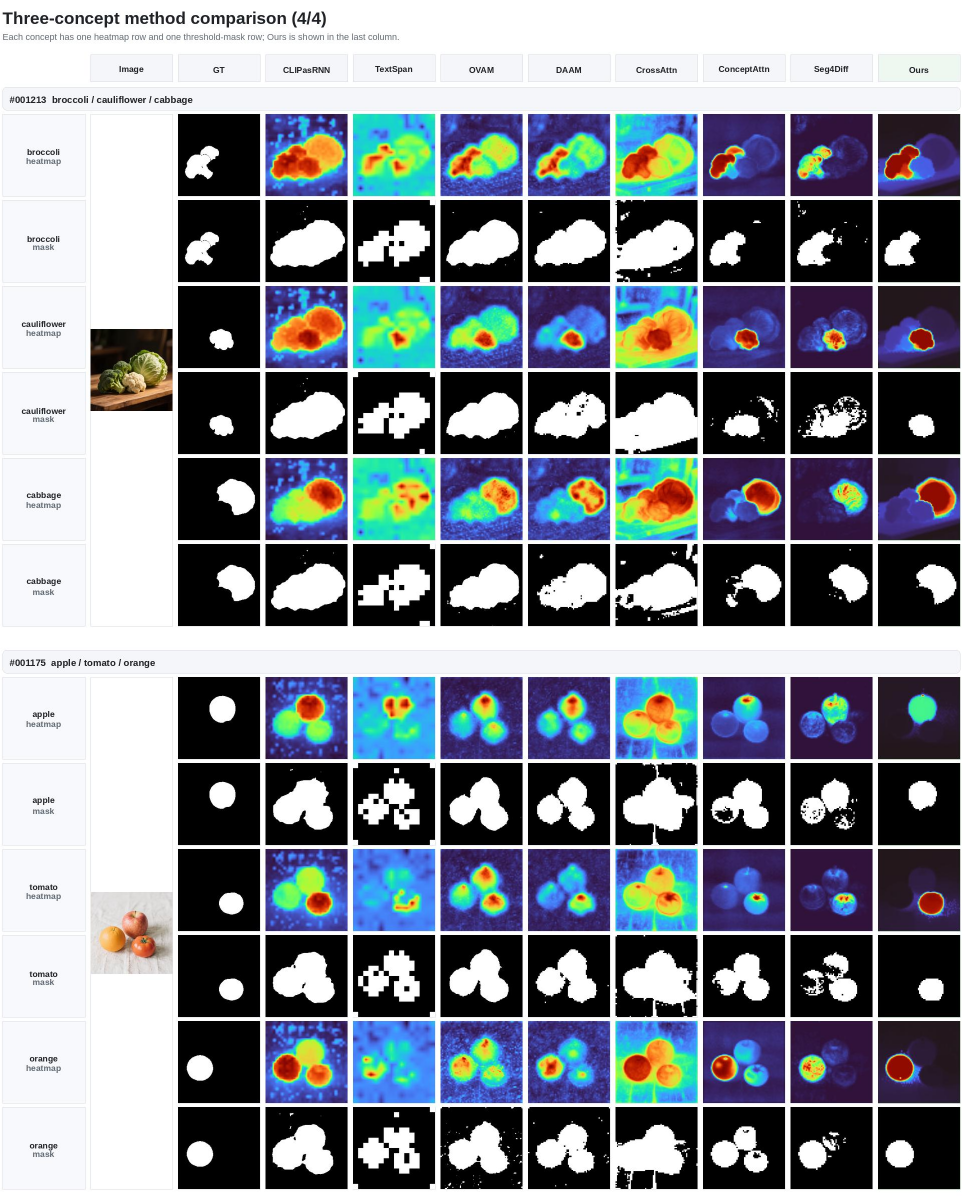}
\caption{
Additional qualitative comparisons on three-concept samples from the Multi-Concept Confusion Dataset. Page 4 of 4.
}
\label{fig:app_three_concept_4}
\end{figure*}
\clearpage

\subsection{Additional layer selection analysis}
\label{sec:layers_analysis}

\paragraph{Single-layer sensitivity.}
\Cref{tab:appendix_single_layer_sd3,tab:appendix_single_layer_sd35} report numerical results when using a single image-to-image attention layer for graph propagation. Different layers show distinct trade-offs between target-object coverage and leakage suppression. For SD3, L9 is the strongest single layer, achieving the best mIoU$_\mathrm{fg}$ and Acc$_\mathrm{fg}$ while also producing the lowest NAR. Deeper layers such as L18 provide complementary behavior, with strong mAP$_\mathrm{fg}$ but weaker foreground accuracy, motivating the use of layer fusion.

For SD3.5-Large, strong single-layer performance appears in later layers, especially L20, L23--L25, L31, and L37. L25 achieves the highest single-layer mIoU$_\mathrm{fg}$, while L37 gives the lowest NAR, indicating stronger suppression of non-target activation but not necessarily the best target coverage. This confirms that optimizing a single metric is insufficient for layer selection. The selected two-layer configurations in the main experiments therefore aim to combine complementary structural cues across depths, improving object coverage while controlling concept leakage.

\begin{table}[H]
\vspace{-1em}
\centering
\caption{
Layer-pair selection ablation on the Multi-Concept Confusion Dataset. For each backbone, we evaluate all $\binom{8}{2}=28$ two-layer fusion pairs formed from the top-8 single-layer candidates identified by the layer-wise sensitivity analysis. This table supports the layer choices used in the main experiments by comparing different two-layer structural propagation configurations. Rows are sorted by mIoU$_\mathrm{fg}$ within each backbone. Higher is better for mIoU$_\mathrm{fg}$, mAP$_\mathrm{fg}$, and Acc$_\mathrm{fg}$, while lower is better for NAR. The best value of each metric within each backbone is shown in bold.
}
\label{tab:appendix_layer_pair_selection}
\scriptsize
\setlength{\tabcolsep}{4pt}
\resizebox{\linewidth}{!}{
\begin{tabular}{lcccc|lcccc}
\toprule
\multicolumn{5}{c|}{\textbf{SD3.5}} & \multicolumn{5}{c}{\textbf{SD3}} \\
Layers & mIoU$_\mathrm{fg}$ $\uparrow$ & mAP$_\mathrm{fg}$ $\uparrow$ & NAR $\downarrow$ & Acc$_\mathrm{fg}$ $\uparrow$
& Layers & mIoU$_\mathrm{fg}$ $\uparrow$ & mAP$_\mathrm{fg}$ $\uparrow$ & NAR $\downarrow$ & Acc$_\mathrm{fg}$ $\uparrow$ \\
\midrule
L23+L31 & \textbf{68.85} & \textbf{89.38} & 12.21 & 89.97 & L9+L18 & \textbf{64.13} & 87.27 & 16.86 & 86.24 \\
L10+L23 & 67.25 & 87.18 & 12.07 & 87.73 & L9+L10 & 62.96 & 82.99 & 15.71 & \textbf{88.13} \\
L24+L31 & 67.05 & 87.99 & 12.07 & 88.56 & L9+L14 & 62.37 & 87.17 & 18.26 & 85.97 \\
L25+L31 & 66.48 & 87.87 & 11.89 & 87.46 & L5+L9 & 61.61 & 85.92 & 18.47 & 85.06 \\
L20+L23 & 66.16 & 84.31 & 11.55 & 89.32 & L10+L18 & 61.26 & 88.77 & 20.93 & 83.46 \\
L10+L24 & 65.71 & 86.34 & 11.66 & 86.66 & L9+L11 & 60.95 & 82.32 & 18.34 & 85.36 \\
L20+L31 & 65.31 & 87.01 & 14.70 & 85.36 & L10+L13 & 59.35 & 84.17 & 17.58 & 83.35 \\
L23+L25 & 65.21 & 81.57 & 10.26 & 89.90 & L9+L13 & 59.05 & 83.30 & \textbf{15.00} & 83.68 \\
L23+L24 & 64.90 & 79.71 & 10.42 & \textbf{90.90} & L13+L18 & 57.38 & 81.03 & 21.35 & 79.29 \\
L20+L24 & 64.45 & 83.07 & 11.47 & 87.90 & L9+L15 & 57.15 & 83.38 & 16.86 & 81.70 \\
L10+L31 & 63.49 & 86.76 & 16.06 & 82.71 & L10+L14 & 57.06 & \textbf{89.28} & 23.31 & 81.45 \\
L10+L20 & 62.83 & 85.92 & 14.07 & 83.71 & L5+L10 & 56.99 & 87.89 & 23.43 & 80.56 \\
L10+L25 & 62.83 & 86.39 & 11.75 & 84.80 & L5+L18 & 56.93 & 84.58 & 26.32 & 77.16 \\
L24+L25 & 62.69 & 80.52 & 10.27 & 88.07 & L14+L18 & 56.91 & 82.98 & 27.22 & 76.74 \\
L20+L25 & 62.49 & 83.55 & 11.34 & 86.35 & L13+L14 & 56.85 & 80.52 & 22.82 & 78.75 \\
L23+L37 & 62.01 & 86.95 & 9.44 & 82.85 & L10+L11 & 56.66 & 84.73 & 23.36 & 80.94 \\
L23+L28 & 58.75 & 86.10 & 9.98 & 82.79 & L11+L18 & 56.31 & 82.12 & 26.84 & 76.62 \\
L24+L37 & 58.66 & 86.38 & \textbf{9.33} & 81.20 & L5+L13 & 56.18 & 81.96 & 22.01 & 78.74 \\
L31+L37 & 58.40 & 86.91 & 11.11 & 80.60 & L10+L15 & 56.10 & 83.70 & 20.51 & 79.88 \\
L20+L37 & 56.60 & 86.01 & 10.40 & 80.02 & L11+L13 & 55.21 & 78.81 & 22.55 & 78.51 \\
L28+L31 & 56.33 & 86.19 & 12.28 & 80.24 & L15+L18 & 53.24 & 78.51 & 25.11 & 74.98 \\
L24+L28 & 55.83 & 85.23 & 9.93 & 81.28 & L5+L14 & 53.08 & 84.52 & 29.20 & 74.32 \\
L25+L37 & 54.91 & 86.16 & 9.39 & 79.40 & L5+L11 & 52.91 & 82.42 & 28.92 & 74.44 \\
L20+L28 & 53.50 & 84.81 & 11.50 & 79.41 & L14+L15 & 52.19 & 76.91 & 27.16 & 73.78 \\
L25+L28 & 52.67 & 85.19 & 9.95 & 79.71 & L11+L14 & 52.17 & 80.88 & 29.95 & 73.45 \\
L10+L37 & 49.30 & 85.24 & 11.27 & 76.48 & L5+L15 & 51.71 & 79.05 & 26.24 & 73.81 \\
L10+L28 & 49.05 & 85.03 & 12.03 & 76.86 & L11+L15 & 51.21 & 75.71 & 26.80 & 73.86 \\
L28+L37 & 39.29 & 84.76 & 9.58 & 72.23 & L13+L15 & 48.91 & 76.96 & 22.25 & 74.77 \\
\bottomrule
\end{tabular}}
\end{table}

\paragraph{Layer combination selection.}
\Cref{tab:appendix_layer_pair_selection} reports the full two-layer sweep over the top-8 single-layer candidates ranked by mIoU$_\mathrm{fg}$ for each backbone. For SD3, L9+L18 achieves the highest foreground mIoU among all evaluated two-layer pairs, confirming the configuration used in the main experiments. For SD3.5-Large, L23+L31 achieves the best mIoU$_\mathrm{fg}$ and mAP$_\mathrm{fg}$, supporting our use of this pair as the default SD3.5 configuration. The results also show that minimizing NAR alone is insufficient: several pairs with lower NAR under-cover the target object and obtain lower mIoU$_\mathrm{fg}$ or Acc$_\mathrm{fg}$.

We further evaluate higher-order layer fusions by randomly sampling 20 three-layer and 20 four-layer combinations from the same top-performing single layers, with results reported in \Cref{tab:multi_layer_combinations}. For SD3, a few multi-layer variants slightly exceed L9+L18 (64.13 mIoU$_\mathrm{fg}$): the best three-layer configuration L9+L14+L18 reaches 66.64 and the best four-layer L5+L9+L10+L18 reaches 66.80. For SD3.5-Large, however, none of the sampled multi-layer combinations outperform L23+L31 (68.85), with the best three-layer reaching only 66.74 and the best four-layer reaching 61.20. The improvements are inconsistent across backbones and do not justify the additional complexity, so we adopt the simpler two-layer configuration in the main experiments. Overall, these results show that the best layer combinations are not obtained by selecting layers based on a single metric alone, nor by simply adding more layers. Instead, effective fusion requires complementary structural cues across depths that balance target-object coverage and leakage suppression.

\begin{table}[H]
\centering
\caption{
Single-layer sensitivity on the Multi-Concept Confusion Dataset for SD3 Medium. We evaluate graph propagation using one image-to-image attention layer at a time. Higher is better for mIoU$_\mathrm{fg}$, mAP$_\mathrm{fg}$, and Acc$_\mathrm{fg}$, while lower is better for NAR. The layers used in the main two-layer configuration are highlighted in bold.
}
\label{tab:appendix_single_layer_sd3}
\scriptsize
\setlength{\tabcolsep}{5pt}
\renewcommand{\arraystretch}{0.95}
\resizebox{\linewidth}{!}{
\begin{tabular}{lcccc|lcccc}
\toprule
Layer & mIoU$_\mathrm{fg}$ $\uparrow$ & mAP$_\mathrm{fg}$ $\uparrow$ & NAR $\downarrow$ & Acc$_\mathrm{fg}$ $\uparrow$
& Layer & mIoU$_\mathrm{fg}$ $\uparrow$ & mAP$_\mathrm{fg}$ $\uparrow$ & NAR $\downarrow$ & Acc$_\mathrm{fg}$ $\uparrow$ \\
\midrule
L0  & 10.31 & 15.47 & 53.67 & 49.96 & L12 & 20.08 & 65.68 & 40.31 & 63.31 \\
L1  & 14.97 & 16.36 & 39.49 & 78.53 & L13 & 50.05 & 77.72 & 20.33 & 76.39 \\
L2  & 28.54 & 78.69 & 34.96 & 61.58 & L14 & 34.33 & 80.61 & 30.44 & 65.73 \\
L3  & 12.47 & 20.94 & 54.21 & 47.23 & L15 & 42.92 & 72.23 & 26.87 & 68.90 \\
L4  & 20.21 & 67.53 & 46.83 & 53.09 & L16 & 33.25 & 69.88 & 38.02 & 58.11 \\
L5  & 36.06 & 81.33 & 30.14 & 64.22 & L17 & 17.86 & 57.55 & 48.44 & 51.85 \\
L6  & 28.12 & 83.26 & 33.83 & 63.94 & \textbf{L18} & 42.28 & 82.06 & 25.24 & 70.92 \\
L7  & 21.48 & 75.63 & 42.85 & 55.17 & L19 & 18.69 & 67.29 & 43.06 & 58.76 \\
L8  & 27.72 & 85.78 & 31.42 & 67.44 & L20 & 22.38 & 75.75 & 39.58 & 58.57 \\
\textbf{L9}  & \textbf{55.32} & 80.47 & \textbf{13.78} & \textbf{84.32} & L21 & 19.94 & 72.98 & 42.70 & 56.60 \\
L10 & 41.97 & 85.62 & 19.80 & 76.76 & L22 & 16.24 & 50.78 & 45.86 & 59.38 \\
L11 & 39.26 & 74.75 & 30.62 & 66.00 & L23 & 16.70 & 49.67 & 45.41 & 58.70 \\
\bottomrule
\end{tabular}
}
\end{table}

\begin{table}[H]
\vspace{-1em}
\centering
\caption{
Single-layer sensitivity on the Multi-Concept Confusion Dataset for SD3.5 Large. We evaluate graph propagation using one image-to-image attention layer at a time. Higher is better for mIoU$_\mathrm{fg}$, mAP$_\mathrm{fg}$, and Acc$_\mathrm{fg}$, while lower is better for NAR. The layers used in the main two-layer configuration are highlighted in bold.
}
\label{tab:appendix_single_layer_sd35}
\scriptsize
\setlength{\tabcolsep}{4pt}
\renewcommand{\arraystretch}{0.9}
\resizebox{\linewidth}{!}{
\begin{tabular}{lcccc|lcccc}
\toprule
Layer & mIoU$_\mathrm{fg}$ $\uparrow$ & mAP$_\mathrm{fg}$ $\uparrow$ & NAR $\downarrow$ & Acc$_\mathrm{fg}$ $\uparrow$
& Layer & mIoU$_\mathrm{fg}$ $\uparrow$ & mAP$_\mathrm{fg}$ $\uparrow$ & NAR $\downarrow$ & Acc$_\mathrm{fg}$ $\uparrow$ \\
\midrule
L0  & 14.29 & 15.54 & 42.86 & 74.92 & L19 & 17.75 & 73.81 & 42.13 & 61.50 \\
L1  & 16.13 & 26.58 & 51.06 & 49.96 & L20 & 57.96 & 81.96 & 14.34 & 83.77 \\
L2  & 25.83 & 60.89 & 40.08 & 59.11 & L21 & 29.15 & 76.48 & 39.33 & 55.99 \\
L3  & 25.84 & 46.35 & 34.00 & 57.35 & L22 & 46.40 & 83.47 & 16.07 & 82.10 \\
L4  & 13.99 & 20.74 & 51.85 & 49.11 & \textbf{L23} & 57.00 & 79.91 & 10.75 & \textbf{88.82} \\
L5  & 26.43 & 48.22 & 37.09 & 54.74 & L24 & 58.34 & 78.14 & 10.68 & 88.81 \\
L6  & 35.86 & 63.41 & 33.88 & 61.92 & L25 & \textbf{61.40} & 80.95 & 10.32 & 88.43 \\
L7  & 38.66 & 77.36 & 26.07 & 70.49 & L26 & 48.96 & 83.78 & 18.78 & 78.39 \\
L8  & 35.95 & 77.65 & 29.06 & 66.70 & L27 & 41.74 & 78.29 & 25.41 & 70.70 \\
L9  & 15.13 & 22.31 & 53.55 & 49.30 & L28 & 56.28 & 82.97 & 10.73 & 81.51 \\
L10 & 53.71 & 81.78 & 17.63 & 77.84 & L29 & 41.27 & 85.12 & 25.64 & 68.76 \\
L11 & 24.37 & 75.83 & 39.12 & 59.51 & L30 & 27.81 & 79.25 & 36.95 & 58.71 \\
L12 & 19.20 & 63.30 & 47.24 & 53.60 & \textbf{L31} & 53.44 & 86.36 & 17.59 & 79.19 \\
L13 & 51.25 & 85.10 & 16.87 & 79.56 & L32 & 22.06 & 66.74 & 45.21 & 52.37 \\
L14 & 24.56 & 76.68 & 41.37 & 55.45 & L33 & 21.87 & 76.37 & 40.46 & 58.40 \\
L15 & 42.61 & \textbf{88.58} & 20.43 & 76.07 & L34 & 27.07 & 78.49 & 38.40 & 57.13 \\
L16 & 52.25 & 85.34 & 16.29 & 79.74 & L35 & 33.88 & 64.38 & 12.35 & 68.31 \\
L17 & 38.11 & 84.14 & 28.65 & 65.71 & L36 & 16.49 & 50.80 & 45.36 & 60.18 \\
L18 & 26.41 & 86.42 & 32.59 & 65.54 & L37 & 60.96 & 83.74 & \textbf{9.40} & 82.49 \\
\bottomrule
\end{tabular}
}
\end{table}

\begin{table}[t]
\centering
\caption{Three-layer and four-layer combination ablation on the Multi-Concept Confusion Dataset. We randomly sample 20 three-layer and 20 four-layer combinations from the top-performing single layers identified in the layer-wise sensitivity analysis. Rows are sorted by $\mathrm{mIoU}_{\mathrm{fg}}$ within each subsection.}
\label{tab:multi_layer_combinations}
\small
\setlength{\tabcolsep}{4pt}
\resizebox{\linewidth}{!}{
\begin{tabular}{lcccc|lcccc}
\toprule
\multicolumn{5}{c|}{\textbf{SD3.5}} & \multicolumn{5}{c}{\textbf{SD3}} \\
Layers & mIoU$_{\mathrm{fg}}\uparrow$ & mAP$_{\mathrm{fg}}\uparrow$ & NAR$\downarrow$ & Acc$_{\mathrm{fg}}\uparrow$ & Layers & mIoU$_{\mathrm{fg}}\uparrow$ & mAP$_{\mathrm{fg}}\uparrow$ & NAR$\downarrow$ & Acc$_{\mathrm{fg}}\uparrow$ \\
\midrule
\multicolumn{10}{c}{\textit{Three-layer combinations}} \\
\midrule
L23+L24+L31 & 66.74 & 87.93 & 11.24 & 88.09 & L9+L14+L18  & 66.64 & 88.16 & 19.79 & 85.88 \\
L10+L23+L31 & 66.20 & 88.83 & 12.80 & 85.19 & L5+L9+L18   & 65.66 & 87.74 & 19.61 & 85.46 \\
L10+L20+L23 & 63.36 & 87.17 & 12.07 & 84.40 & L5+L9+L10   & 64.82 & 86.45 & 18.51 & 87.27 \\
L10+L23+L25 & 62.00 & 86.88 & 10.98 & 84.30 & L9+L11+L18  & 64.14 & 86.49 & 19.69 & 85.42 \\
L20+L25+L31 & 61.98 & 87.28 & 12.19 & 84.01 & L10+L11+L14 & 61.31 & 87.54 & 25.03 & 82.66 \\
L10+L20+L31 & 61.52 & 87.51 & 14.40 & 82.24 & L10+L13+L18 & 61.31 & 85.70 & 19.22 & 82.90 \\
L10+L24+L25 & 59.29 & 86.21 & 10.86 & 82.89 & L5+L10+L13  & 60.85 & 85.64 & 20.08 & 83.00 \\
L10+L20+L25 & 58.48 & 86.50 & 11.89 & 82.12 & L9+L13+L18  & 60.02 & 85.33 & 16.96 & 82.70 \\
L23+L31+L37 & 57.59 & 88.48 & 10.48 & 80.21 & L5+L9+L13   & 59.77 & 85.13 & 17.57 & 83.01 \\
L23+L24+L28 & 54.59 & 85.91 &  9.93 & 80.56 & L5+L13+L14  & 59.22 & 83.08 & 23.75 & 79.98 \\
L24+L28+L31 & 54.02 & 87.33 & 10.98 & 79.34 & L5+L13+L18  & 59.02 & 83.24 & 22.48 & 80.05 \\
L23+L25+L37 & 53.33 & 87.20 &  9.51 & 78.60 & L10+L15+L18 & 58.99 & 85.14 & 21.48 & 80.52 \\
L20+L24+L37 & 52.45 & 86.86 & 10.05 & 78.05 & L5+L9+L15   & 58.32 & 85.05 & 19.42 & 81.38 \\
L25+L31+L37 & 52.02 & 87.76 & 10.39 & 77.67 & L5+L10+L15  & 58.31 & 85.13 & 22.58 & 80.31 \\
L20+L25+L37 & 49.97 & 86.75 & 10.08 & 76.98 & L11+L13+L14 & 57.98 & 80.88 & 24.24 & 79.46 \\
L24+L25+L28 & 49.63 & 85.31 &  9.91 & 78.20 & L5+L11+L13  & 57.54 & 81.94 & 23.50 & 79.69 \\
L10+L20+L37 & 46.13 & 86.52 & 11.36 & 75.07 & L10+L11+L15 & 57.13 & 83.45 & 22.67 & 80.23 \\
L10+L25+L37 & 45.46 & 86.70 & 10.41 & 74.77 & L11+L14+L15 & 54.78 & 78.37 & 27.41 & 75.86 \\
L10+L25+L28 & 45.26 & 86.34 & 10.84 & 75.22 & L9+L13+L15  & 52.22 & 82.78 & 17.19 & 78.88 \\
L20+L28+L37 & 37.58 & 85.92 & 10.18 & 71.44 & L13+L15+L18 & 51.18 & 79.57 & 22.43 & 76.00 \\
\midrule
\multicolumn{10}{c}{\textit{Four-layer combinations}} \\
\midrule
L23+L24+L25+L31 & 61.20 & 86.89 & 10.81 & 84.45 & L5+L9+L10+L18    & 66.80 & 88.15 & 19.25 & 86.46 \\
L10+L20+L23+L24 & 59.03 & 86.83 & 11.26 & 82.21 & L10+L11+L14+L18  & 65.73 & 87.79 & 24.58 & 83.86 \\
L10+L20+L23+L25 & 56.80 & 86.84 & 11.26 & 81.24 & L9+L10+L11+L18   & 65.13 & 87.15 & 19.24 & 86.40 \\
L20+L23+L28+L31 & 52.02 & 87.70 & 11.43 & 78.24 & L5+L9+L11+L18    & 64.89 & 87.09 & 21.30 & 84.67 \\
L20+L23+L24+L37 & 51.00 & 87.41 & 10.02 & 77.36 & L5+L9+L10+L11    & 64.41 & 86.06 & 20.19 & 86.49 \\
L23+L25+L28+L31 & 50.36 & 87.66 & 10.62 & 77.66 & L5+L11+L14+L18   & 63.31 & 84.63 & 27.53 & 80.44 \\
L20+L24+L25+L28 & 46.47 & 85.53 & 10.41 & 76.42 & L5+L10+L13+L18   & 61.44 & 86.32 & 20.83 & 82.38 \\
L10+L20+L23+L37 & 45.18 & 87.54 & 10.80 & 74.65 & L5+L9+L13+L14    & 60.72 & 86.07 & 19.48 & 82.65 \\
L10+L24+L31+L37 & 45.05 & 87.91 & 11.04 & 74.35 & L5+L9+L10+L13    & 60.15 & 86.00 & 17.54 & 83.34 \\
L10+L25+L28+L31 & 44.58 & 87.49 & 11.55 & 74.58 & L9+L10+L14+L15   & 59.98 & 86.23 & 18.96 & 82.69 \\
L10+L23+L25+L37 & 44.42 & 87.50 & 10.19 & 74.29 & L5+L13+L14+L18   & 59.92 & 83.75 & 23.81 & 80.17 \\
L10+L20+L24+L37 & 43.64 & 87.20 & 10.67 & 73.95 & L9+L11+L13+L14   & 59.39 & 84.75 & 19.65 & 82.42 \\
L10+L25+L31+L37 & 43.50 & 87.75 & 11.10 & 73.70 & L5+L11+L13+L14   & 59.03 & 83.00 & 24.71 & 80.00 \\
L10+L24+L25+L28 & 43.43 & 86.54 & 10.44 & 74.39 & L11+L13+L14+L18  & 58.97 & 82.25 & 24.18 & 79.75 \\
L10+L20+L25+L28 & 42.89 & 86.53 & 11.17 & 74.07 & L5+L14+L15+L18   & 57.51 & 82.24 & 26.14 & 77.75 \\
L23+L24+L28+L37 & 37.61 & 87.07 &  9.50 & 71.44 & L5+L11+L15+L18   & 56.51 & 81.84 & 25.85 & 77.55 \\
L24+L25+L28+L37 & 35.31 & 86.63 &  9.49 & 70.53 & L10+L13+L14+L15  & 53.81 & 83.71 & 21.07 & 78.65 \\
L10+L28+L31+L37 & 33.98 & 86.79 & 11.27 & 69.57 & L10+L11+L13+L15  & 52.38 & 82.78 & 20.91 & 78.30 \\
L10+L24+L28+L37 & 33.21 & 86.68 & 10.11 & 69.35 & L5+L13+L14+L15   & 52.01 & 81.20 & 23.84 & 76.46 \\
L10+L25+L28+L37 & 32.30 & 86.56 & 10.19 & 69.01 & L5+L13+L15+L18   & 52.00 & 81.54 & 22.93 & 76.53 \\
\bottomrule
\end{tabular}
}
\end{table}

\subsection{Visualization of the effect of \texorpdfstring{$W_{\cos}$}{Wcos}}
\label{app:wcos_visualization}

\begin{figure}[t]
    \centering
    \includegraphics[width=\linewidth]{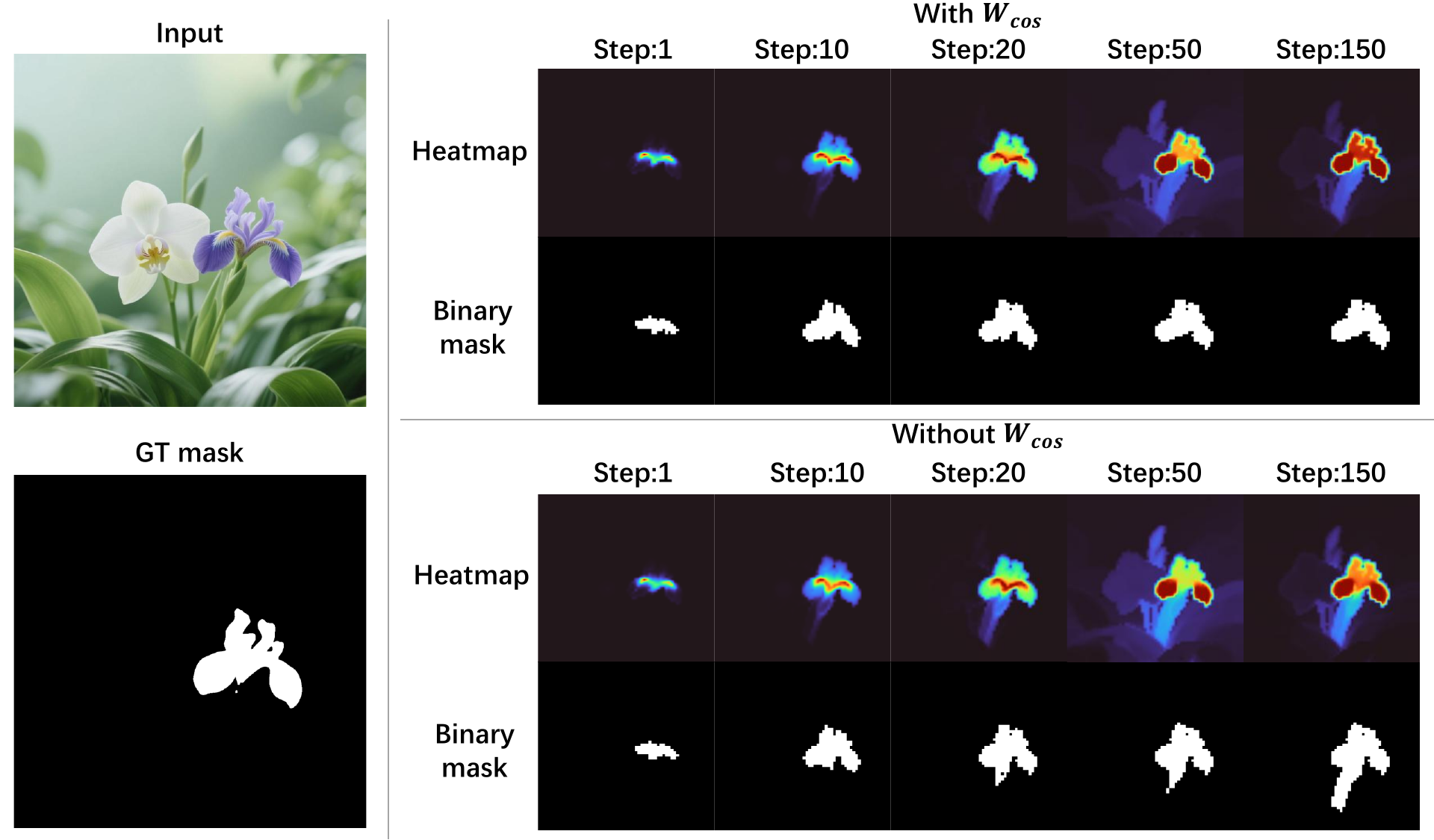}
    \caption{
    Visualization of graph propagation with and without $W_{\cos}$.
    $W_{\cos}$ improves within-object propagation, while the attention gate helps suppress
    leakage to nearby non-target regions. Example 1 of 3.
    }
    \label{fig:wcos_vis_1}
\end{figure}

\begin{figure}[t]
    \centering
    \includegraphics[width=\linewidth]{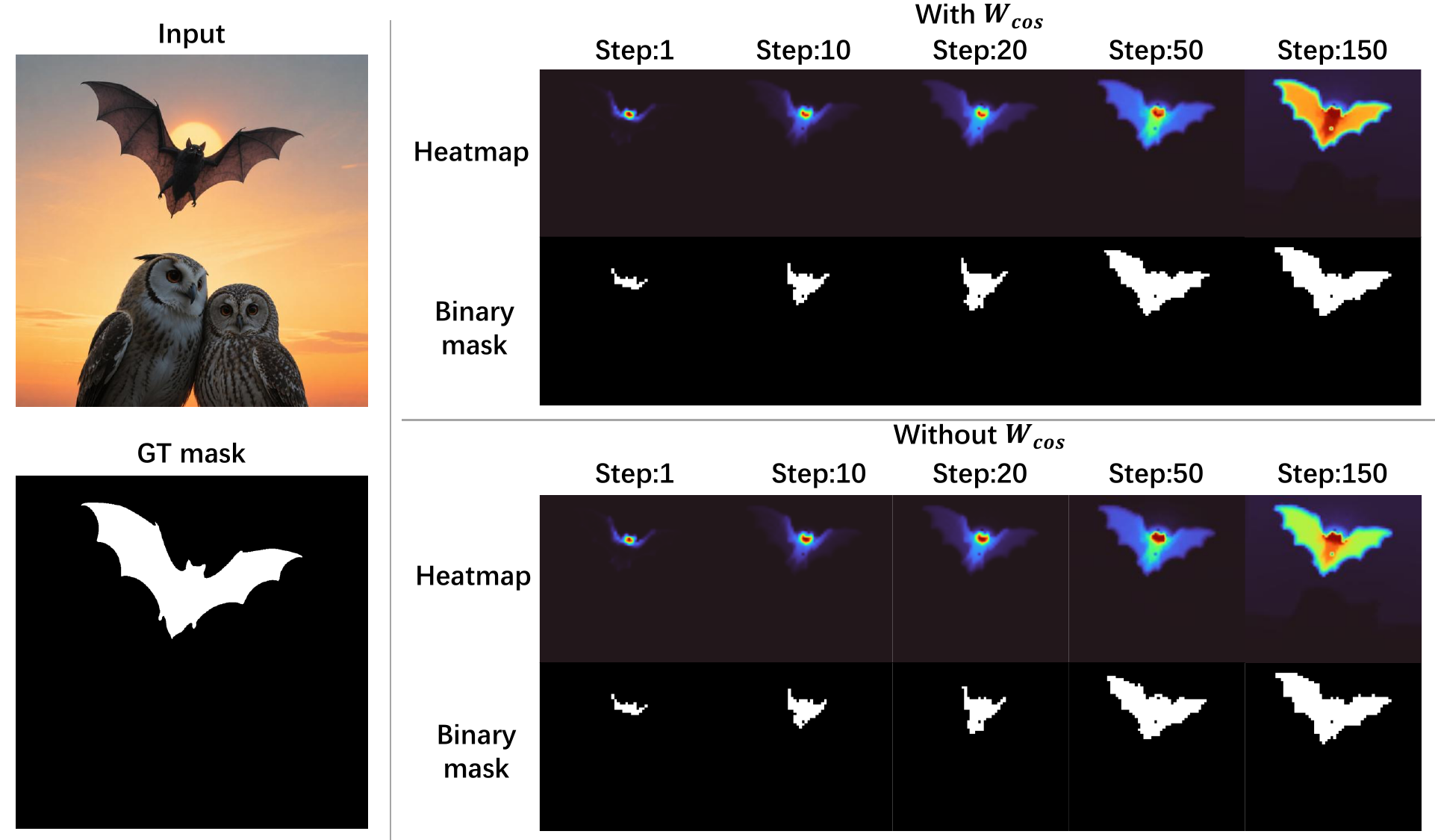}
    \caption{
    Visualization of graph propagation with and without $W_{\cos}$.
    $W_{\cos}$ improves within-object propagation, while the attention gate helps suppress
    leakage to nearby non-target regions. Example 2 of 3.
    }
    \label{fig:wcos_vis_2}
\end{figure}

\begin{figure}[t]
    \centering
    \includegraphics[width=\linewidth]{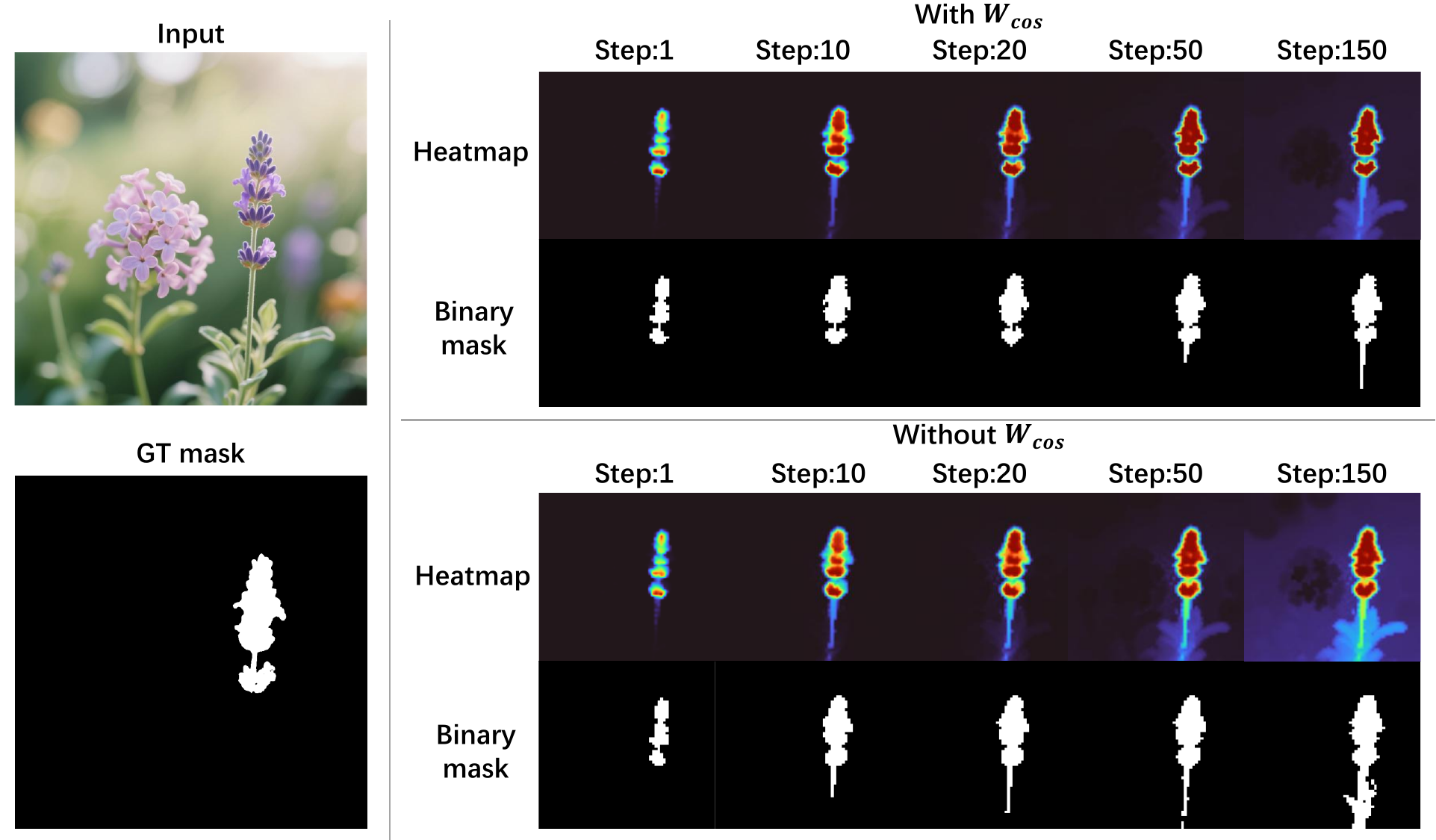}
    \caption{
    Visualization of graph propagation with and without $W_{\cos}$.
    $W_{\cos}$ improves within-object propagation, while the attention gate helps suppress
    leakage to nearby non-target regions. Example 3 of 3.
    }
    \label{fig:wcos_vis_3}
\end{figure}

We further provide qualitative ablations to illustrate the role of the output-space affinity $W_{\cos}$ in the proposed graph propagation process. As shown in \Cref{fig:wcos_vis_1,fig:wcos_vis_2,fig:wcos_vis_3}, we compare the propagation trajectories with and without $W_{\cos}$ at different propagation steps. These examples contain target objects and visually related non-target regions, making them suitable for analyzing both
within-object propagation and cross-object leakage.

As shown in \Cref{fig:wcos_vis_2}, with $W_{\cos}$, the heat response spreads more smoothly over the entire target object as the propagation step increases. This is because $W_{\cos}$ is computed from the self-attention output features and provides dense long-range affinity between patches belonging to the same object. Consequently, patches on different parts of the target object remain strongly connected, allowing the one-hot anchor response to propagate from the initial high-confidence seed to the full object region. The resulting heatmaps are more spatially uniform inside the target object, and the corresponding binary masks recover the object shape more consistently. In contrast, removing $W_{\cos}$ weakens the dense within-object connectivity. Although the row-wise attention gate can still constrain propagation to structurally related regions, the propagation becomes less effective at transferring activation across distant parts of the same object.

As reported in \Cref{tab:property3_affinity_gating}, $W_{\cos}$ assigns high affinity to same-object patches, but it can also connect visually confusable different objects. In our full model, the attention gate suppresses such cross-object and foreground-background connections, thereby enlarging the relative similarity gap between the target object and other objects. This effect is also reflected in the visual comparisons in \Cref{fig:wcos_vis_1} and \Cref{fig:wcos_vis_3}. With $W_{\cos}$, the target-object heatmap is brighter and more coherent inside the object, while the surrounding non-target regions remain darker. Without $W_{\cos}$, the response around the target is less well confined and is more prone to leakage. These visualizations support the quantitative ablation results and show that $W_{\cos}$ mainly improves object-level response completeness, while the attention gate prevents this dense propagation from spreading to visually similar non-target objects.

\end{document}